\documentclass[letterpaper]{article} 
\usepackage{aaai25}  
\usepackage{times}  
\usepackage{helvet}  
\usepackage{courier}  
\usepackage[hyphens]{url}  
\usepackage{graphicx} 
\usepackage{natbib}  
\usepackage{caption} 
\usepackage{tcolorbox}
\usepackage{algorithm}
\usepackage{algorithmic}

\usepackage{newfloat}
\usepackage{listings}
\DeclareCaptionStyle{ruled}{labelfont=normalfont,labelsep=colon,strut=off} 
\floatstyle{ruled}
\newfloat{listing}{tb}{lst}{}
\floatname{listing}{Listing}
\nocopyright 
\usepackage{graphicx,subfigure}

\usepackage[utf8]{inputenc} 
\usepackage[T1]{fontenc}    
\usepackage{url}            
\usepackage{booktabs}       
\usepackage{amsfonts}       
\usepackage{nicefrac}       
\usepackage{microtype}      
\usepackage{xcolor}         

\usepackage{enumitem}
\usepackage{amsfonts}
\usepackage{multirow}
\usepackage{lipsum} 
\usepackage{multicol}
\usepackage{arydshln}
\usepackage{capt-of}
\usepackage{mathrsfs}
\usepackage{array}
\usepackage{bm}
\usepackage{color}
\usepackage{colortbl}
\usepackage{epsfig}
\usepackage{float}
\usepackage{lineno}
\usepackage{indentfirst}
\usepackage{makecell}
\usepackage[ragged]{footmisc}
\usepackage{scrextend}
\usepackage{xurl}
\usepackage{amsmath}
\usepackage{amssymb}
\usepackage{mathtools}
\usepackage{amsthm}
\usepackage{soul}
\usepackage{graphicx,lipsum}
 
\title{Enhance Hyperbolic Representation Learning via Second-order Pooling}
\author{
    Kun Song\textsuperscript{\rm 1}\equalcontrib,
    Ruben Solozabal\textsuperscript{\rm 1}\equalcontrib,
    Li hao\textsuperscript{\rm 2},
    Lu Ren\textsuperscript{\rm 2},
    Moloud Abdar\textsuperscript{\rm 3},
    Qing Li\textsuperscript{\rm 1}, \\
    Fakhri Karray\textsuperscript{\rm 1}, and 
    Martin Takáč\textsuperscript{\rm 1}
}
\affiliations{
    \textsuperscript{\rm 1}Mohamed bin Zayed University of Artificial Intelligence (MBZUAI), Abu Dhabi, UAE\\
    \textsuperscript{\rm 2}Anhui University, Anhui Province, China\\
    \textsuperscript{\rm 3} Deakin University, Australia\\

}

\usepackage{bibentry}

\begin{document}

\maketitle

\begin{abstract}

 Hyperbolic representation learning is well known for its ability to capture hierarchical information. However, the distance between samples from different levels of hierarchical classes can be required large. We reveal that the hyperbolic discriminant objective forces the backbone to capture this hierarchical information, which may inevitably increase the Lipschitz constant of the backbone. This can hinder the full utilization of the backbone's generalization ability. To address this issue, we introduce second-order pooling into hyperbolic representation learning, as it naturally increases the distance between samples without compromising the generalization ability of the input features. In this way, the Lipschitz constant of the backbone does not necessarily need to be large. However, current off-the-shelf low-dimensional bilinear pooling methods cannot be directly employed in hyperbolic representation learning because they inevitably reduce the distance expansion capability. To solve this problem, we propose a kernel approximation regularization, which enables the low-dimensional bilinear features to approximate the kernel function well in low-dimensional space. Finally, we conduct extensive experiments on graph-structured datasets to demonstrate the effectiveness of the proposed method.

\end{abstract}

%


\section{Introduction}
Many datasets contain hierarchical information. For example, in datasets like Cifar~\cite{krizhevsky2009learning} and fine-grained image datasets~\cite{WahCUB_200_2011}, and graph-structured data (e.g., drug molecules, social media networks), categories belonging to superclasses that can be further divided into subclasses, which is a typical hierarchical structure. Thus, inducing the right bias to capture this hierarchy is key to enhancing generalization in machine learning models \cite{nickel2018learning,peng2021hyperbolic}.\\
\indent As hyperbolic space can present hierarchical information \cite{peng2021hyperbolic,nickel2018learning} well, many representation learning techniques based on hyperbolic distance have been proposed, such as the various hyperbolic neural networks \cite{chamberlain2017neural,peng2021hyperbolic,chami2019hyperbolic} etc. Besides those deep hyperbolic structures, there is a routine of approaches depicting hierarchical information by simply using the hyperbolic discriminant objective to train the traditional backbones \cite{ermolov2022hyperbolic,ge2023hyperbolic,zhang2022hyperbolic}. Compared with the deep hyperbolic structures, those methods have fewer hyperparameters to tune and thus are easy to train \cite{yang2023hyperbolic}.\\
\indent The key step of performing hyperbolic objectives on traditional backbones is to project the extracted features into the hyperbolic space via the exponential map. Thus, the features extracted by the backbone present two different distributions: one in Euclidean space (before the exponential map) and the other in the hyperbolic space. In this paper, we find that the hyperbolic distance $d_c^{h}(\textbf{z}_i,\textbf{z}_j)$ ($\textbf{z}_i = \exp(\textbf{x}_i)$) can be upper-bounded by a linear expansion of the Euclidean distance $d^{e}(\textbf{x}_i,\textbf{x}_j)$, i.e., $d_{c}^{h}(\textbf{x}_i,\textbf{x}_j) < Kd^{e}(\textbf{x}_i,\textbf{x}_j)$ where $K>0$ is a constant. The linear upper bound implies that a large hyperbolic distance corresponds to a large Euclidean distance, and the exponential map minimally changes the distribution of the samples. Specifically, we can claim that the hyperbolic geometry-based discriminant analysis forces the features extracted by the backbone to present a hierarchical structure.\\
\begin{figure}[bt]
    \centering
    \footnotesize
    \includegraphics[width=\linewidth]{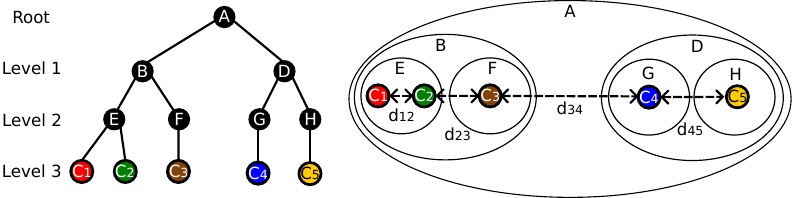}
    \caption{The hierarchical representation of data is retrieved using distances in representation learning.}
    \label{fig:tree}
    \vspace{-3pt}
\end{figure}
%
\begin{figure*}[t]
    \centering
    \footnotesize
  \includegraphics[width=0.7\textwidth]{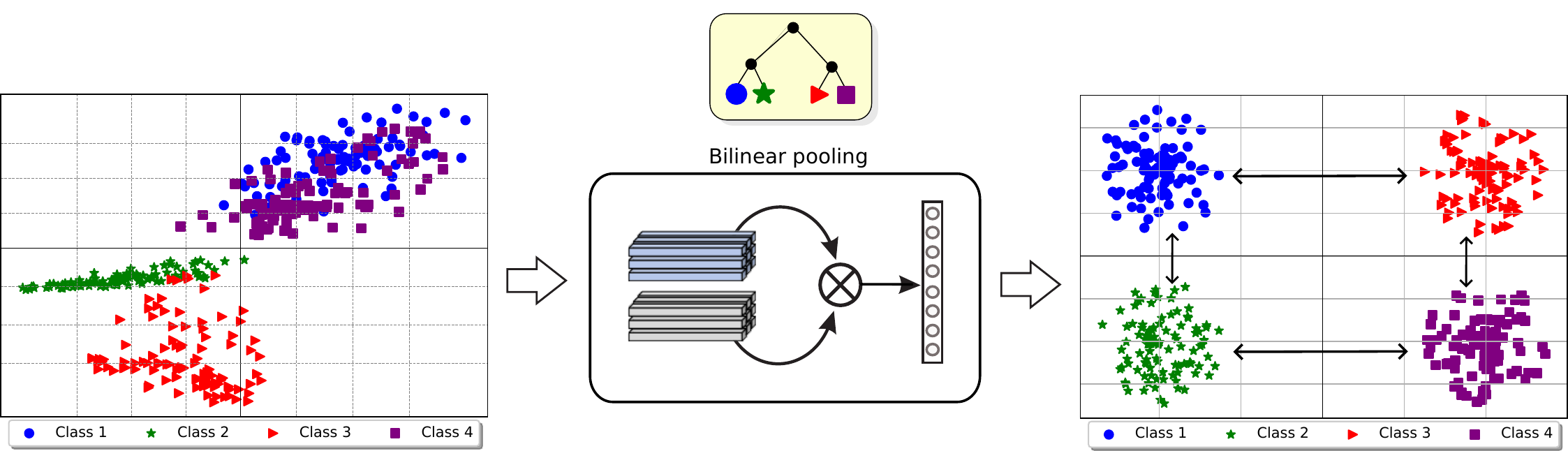}
    \caption{Demonstration of how hierarchical information is enhanced using bilinear pooling. Class 1 and Class 2 belong to the same subgroup, whereas Class 3 and Class 4 belong to another subgroup. On the left, the hierarchical information in the projection is not significant. On the right, after bilinear pooling, the hierarchy in the data representation is observed as clusters of the same subgroup become closer. The exact procedure to compute the inverse bilinear pooling is presented in Appendix A.}
    \label{figure_inverser_bilinear_pooling}
\end{figure*}

\indent However, capturing hierarchical information inevitably leads to large distances among samples belonging to different superclasses. As seen from Figure~\ref{fig:tree}, C1 and C2 are separated in the level 2-superclass (E) with a distance of $d_{12}$ between them. Classes C2 and C3 belong to different super-classes, which are separated in the level one-superclass (B). Thus, their distance is $d_{23}$. We can find $d_{23} > d_{12}$. Because C4 and C3 are separated in the root level-superclass, the distance $d_{34} > d_{23}+d_{12}$. In this way, we can claim that the level of the superclass for separation is closer to the root, the distance between samples considered will be larger. Moreover, the distance will increase exponentially when the depth of the hierarchical structure is large \cite{dhingra2018embedding,chamberlain2017neural}.\\
\indent The Lipschitz constant of a function $f_{\theta}(\textbf{I}_i)$ is $\sup_{\forall \textbf{I}_i,\textbf{I}_j}\frac{\Vert f_{\theta}(\textbf{I}_i) - f_{\theta}(\textbf{I}_j)\Vert_2}{\Vert \textbf{I}_i - \textbf{I}_j\Vert_2}$. Thus, the large distance between extracted features may require a large Lipschitz constant of the backbone. This leads to two significant problems: (1) to enable the backbone to fit a function with a large Lipschitz constant, more layers may be required \cite{song2021adaptive}. For some deep models, such as graph neural networks, how to effectively train a deeper model is still an open problem \cite{liu2021graph,zhu2020bilinear}; (2) A trained backbone with a large Lipschitz constant may suffer from a loss of generalization ability, as a large Lipschitz constant indicates reduced smoothness in the backbone's landscape \cite{fazlyab2019efficient,fazlyab2024certified,krishnan2020lipschitz}. Given the two problems arising from the requirements of hierarchical information extraction, we pose the following question:\\\vspace{-5pt}\\
\emph{Can we design a deep structure that captures hierarchical information without requiring a large Lipschitz constant of the backbone?}\\\vspace{-5pt}\\
\indent To address this question, we introduce bilinear pooling into hyperbolic representation learning to help capture hierarchical information. The reason why we adopt bilinear pooling is that bilinear pooling non-linearly increases distances among its inputted features without compromising their generalization ability. With the help of bilinear pooling, the distances among the backbone's features are not required to be very large. As a result, the Lipschitz constant of the backbone can remain small. As illustrated in Figure~\ref{figure_inverser_bilinear_pooling}, the hierarchical structure in bilinear pooling does not require its input features to present a hierarchical structure.\\
\indent Similar to implementing traditional bilinear pooling approaches, integrating bilinear pooling with hyperbolic representation learning requires reducing the dimension of bilinear features. Although many off-the-shelf low-dimensional bilinear pooling methods are available, they cannot be directly adopted in the hyperbolic representation learning paradigm. We take the low-rank bilinear pooling \cite{gao2020revisiting,kim2016hadamard} and random Maclaurin bilinear pooling ~\cite{gao2016compact,pham2013fast} as examples. Low-rank bilinear pooling can produce low-dimensional bilinear features, but it loses the ability to expand the distances of input features. Random Maclaurin projection can keep that ability of distance expansion, but its performance is not stable, requiring the high dimension of its outputs \cite{yu2021fast}.\\
\indent To address the above problem, we propose a regularization term named kernel approximation regularization ($\mathbb{KR}$), which lets the inner product of the low-dimensional bilinear features approximate a given kernel function. The benefits of $\mathbb{KR}$ consist of two aspects: (1) the extracted low-dimensional bilinear features do not lose the ability to enlarge the distance, and its performance is stable; (2) we can control the non-linearity of learned low-dimensional features by selecting the appropriate kernel functions. At last, we conduct experiments on graph-structure data to demonstrate the effectiveness of the proposed methods.

\section{Preliminaries and Notations}

\subsection{Hyperbolic Geometry}
There are several well-studied isometric models that endow Euclidean space with a hyperbolic metric, such as the Poincaré model and the Lorentz model, which are equivalent \cite{nickel2018learning}. For better understanding, we discuss hyperbolic geometry using the Poincaré model because of its explicit geometrical meaning.\\
\indent The Poincaré model is on a $n$-dimensional sphere, which is defined as a $c$ curvature manifold $D^n_c = \{\textbf{z} \in \mathbb{R}^n: c\Vert \textbf{z}\Vert_2 < 1\}$, with the Riemannian metric $g_c^{\mathbb{D}} = \lambda_c^2(\textbf{z})\cdot g^{E}$, in which $\lambda_{\textbf{z}} = \frac{2}{1-c\Vert \textbf{z}\Vert^2}$ is the conformal factor, and $g^E = \textbf{I}_E$ is the Euclidean metric tensor. Using the defined distance, hyperbolic discriminant analysis on a set of data can be made.\\
\indent The geometry of the Poincaré model can be described using the Möbius gyrovector space as well, which induces a new type of hyperbolic distance widely used in deep learning \cite{}. Firstly, we introduce the Möbius addition for $\textbf{z}_i$, $\textbf{z}_j \in \mathbb{D}^{n}_c$ defined as follows:
\begin{equation}
\label{equation_meq}
\textbf{z}_i\oplus_c \textbf{z}_j =\! \frac{(1\!\!+\!\!2c\langle\textbf{z}_i,\textbf{z}_j\rangle \!+\! c\Vert \textbf{z}_j\Vert_2^2)\textbf{z}_i + (1\!-\! c\Vert \textbf{z}_i\Vert_2^2)\textbf{z}_j}{1+ 2c\langle\textbf{z}_i,\textbf{z}_j\rangle + c^2\Vert \textbf{z}_i \Vert_2^2\Vert \textbf{z}_j\Vert_2^2}.
\end{equation}
The hyperbolic distance between two samples on $\mathbb{D}_c^n$ is:
\begin{equation}
\label{dist_hyperb}
d_c^{h}(\textbf{z}_i,\textbf{z}_j) = \frac{2}{\sqrt{c}} \tanh^{-1}(\sqrt{c}\Vert -\textbf{z}_i \oplus_c\textbf{z}_j\Vert_2).
\end{equation}
For a point $\textbf{z}\in D^{n}_c$, the tangent space at $\textbf{z}$, denoted by $T_z\mathbb{D}_c^n$, is a Euclidean space, which contains the tangent vector with all possible directions at $\textbf{z}$. The exponential map provides a way to project a point $\textbf{x}_i\in T_z\mathbb{D}_c^n$ to the Poincare ball as follows:
\begin{equation}
\label{exp_fun1}
    \textbf{z}_i=\exp_{\textbf{z}}(\textbf{x}_i) = \textbf{z} \oplus_c (\tanh(\sqrt{c}\frac{\Vert \textbf{x}_i \Vert_2}{2})\frac{\textbf{x}_i}{\sqrt{c}\Vert \textbf{x}_i\Vert_2}).
\end{equation}
The inverse process is termed as the logarithm map, which projects a point $\textbf{z}_i \in \mathbb{D}_c^n$, to the tangent plane of $\textbf{x}_i$, given as:
\begin{equation}
\label{logth}
\begin{split}
    \textbf{x}_i & =\log_{\textbf{z}}(\textbf{z}_i)\\
   & = \frac{2}{\sqrt{c}}\tanh^{-1}(\sqrt{c}\Vert \!-\!\textbf{z} \oplus_c \textbf{z}_i \Vert_2)\frac{-\textbf{z}\oplus_c \textbf{z}_i}{\Vert \!\!-\!\!\textbf{z} \oplus_c\! \textbf{z}_i\Vert_2}.
\end{split}
\end{equation}

\begin{figure*}[t]
    \centering
    \footnotesize
    \includegraphics[width=0.75\textwidth]{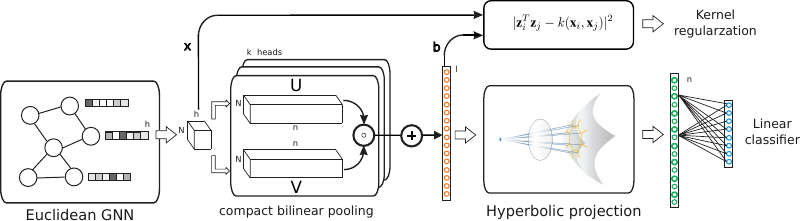}
    \caption{Network architecture. An Euclidean GNN backbone is used for node feature extraction. For each node, bilinear features are expanded and further projected into hyperbolic space. The proposed kernel-based regularization is applied to obtain compact bilinear features.}
    \label{fig:architecuture}
\end{figure*}
\subsection{Deep Hyperbolic Representation Learning}
\indent Given a set of samples $\mathcal{I} = \{(\textbf{I}_i,y_i)\}_{i=1}^N$, where $\textbf{I}_i$ is the $i$-th sample (e.g., images or graphs) and $y_i$ is the corresponding label. A deep neural network $\textbf{x}_i=f_{\theta}(\textbf{I}_i)$ extracts the feature of the sample $\textbf{I}_i$. Then, $\textbf{x}_i$ is projected to the hyperbolic space via the exponential map $\textbf{z}_i= \exp_{\textbf{0}}(\textbf{x}_i)$ defined by Eq.(\ref{exp_fun1}). Thus, the distances among those samples $\{\textbf{z}_i\}_{i=1}^N$ in the hyperbolic space can be calculated via Eq.(\ref{dist_hyperb}). In this way, the loss function based on the hyperbolic distance is calculated as:
\begin{equation}
    J(\theta,\mathcal{I}) = \sum_{i=1}^N \frac{\exp(d_{c}(\textbf{z}_i,\textbf{w}_{y_i}))}{\sum_{k=1}^C \exp(d_{c}(\textbf{z}_i,\textbf{w}_{k}))},
\end{equation}
where $\textbf{w}_k$ is the proxy of the $k$-th class. In this way, the discriminant analysis is performed in the hyperbolic space.\\
\indent In the following content, we want to explore how the hyperbolic distance-based objective affects the training of the backbone.
\begin{tcolorbox}
\textbf{Theorem 1.} Given two samples $\textbf{x}_i$ and $\textbf{x}_j$ in the tangent space of $\textbf{0}$, Eq.(\ref{exp_fun1}) projects them into the hyperbolic space denoted by $\textbf{z}_i$ and $\textbf{z}_j$, respectively, whose hyperbolic distance is $d_{c}^h(\textbf{z}_i,\textbf{z}_j)$. The relation with the distances satisfy:
\begin{equation}
\begin{split}
&d_{c}^h(\textbf{z}_i,\textbf{z}_j) \approx  \sqrt{c}(\Vert \textbf{x}_i\Vert_2 + \Vert \textbf{x}_j \Vert_2), \quad \textbf{x}_i^T\textbf{x}_j = 0\\
&d_{c}^h(\textbf{z}_i,\textbf{z}_j) = \sqrt{c} \Vert \textbf{x}_i - \textbf{x}_j\Vert_2, \!\quad\! \textbf{x}_i = k \textbf{x}_j, k \in \mathbb{R}\\
\end{split}
\end{equation}
\end{tcolorbox}
\indent The proof is attached in the appendix.\\\vspace{-5pt}\\
\textbf{Remark 1.} \emph{From theorem 1, we claim that the hyperbolic distance is bounded by a linear expansion of the Euclidean distance.}
\\\vspace{-5pt}\\
\indent Let us consider the case $\textbf{x}_i^T\textbf{x}_j = 0$. Because $2\sqrt{a^2+b^2} \geq \vert a\vert + \vert b \vert$, there is $d_{c}(\textbf{z}_i,\textbf{z}_j) \approx \sqrt{c}(\Vert \textbf{x}_i\Vert_2 + \Vert \textbf{x}_j \Vert_2) \leq 2\sqrt{c} \Vert \textbf{x}_i -\textbf{x}_j \Vert_2$.\\
\indent Let us discuss the case $\textbf{x}_i^T\textbf{x}_j \neq 0$. Because Eq.(\ref{exp_fun1}) and Eq.(\ref{logth}) are two mutually inverse functions, each point in the tangent space has a map in the hyperbolic space, and vice versa. Thus, given two samples $\textbf{x}_i$ and $\textbf{x}_j$, there is a function between the value of the Euclidean distance $d^{e}(\textbf{x}_i,\textbf{x}_j)$ and the value of the hyperbolic distance $d_{c}^h(\textbf{z}_i,\textbf{z}_j)$, where $\textbf{z}_i = \exp_{\textbf{0}}(\textbf{x}_i)$. Thus, by dividing the two distances, we can obtain a new function denoted by $p(d^{e}(\textbf{x}_i,\textbf{x}_j)) = \frac{d^{e}(\textbf{z}_i,\textbf{z}_j)}{d_{c}^h(\textbf{x}_i,\textbf{x}_j)}$. If we fix the norm $\Vert \textbf{x}_i\Vert_2$ and $\Vert \textbf{x}_j\Vert_2$ and let $\Vert \textbf{x}_i\Vert_2 \neq \Vert \textbf{x}_j\Vert_2$, the function $p(d^{e}(\textbf{x}_i,\textbf{x}_j))$ can be rewritten as $p(\theta)$ where $\theta$ is the angle between $\textbf{x}_i$ and $\textbf{x}_j$. In this way, we have $d_{c}^h(\textbf{z}_i,\textbf{z}_j) = p(\theta) d^e(\textbf{x}_i,\textbf{x}_j)$. From \textbf{Theorem 1}, we know that $p(0) = p(\pi) = \sqrt{c}$ and $p(\pi/2) = p(3\pi/2) \approx 2\sqrt{c}$. Because the maps in Eq.(\ref{exp_fun1}) and Eq.(\ref{logth}) connect two Riemannian manifolds, the maps should be smooth and continuous. Thus, the function $p(\theta)$ is smooth and continuous. Considering $p(0) = p(\pi) = \sqrt{c}$ and $p(\pi/2) = p(3\pi/2) \approx 2\sqrt{c}$, $p(\theta)$ has the maximal value $K$ when $\theta \in [0,2\pi]$. Therefore, we can have $d^h_c(\textbf{z}_i,\textbf{z}_j) < K d^h_c(\textbf{x}_i, \textbf{x}_j)$.\\
\indent  The content in \textbf{remark 1} indicates that if a set of clusters is distributed hierarchically in the hyperbolic space, their origins of the exponential map are also hierarchically distributed (in Euclidean space). This is because the hierarchical distribution inevitably contains samples with a very large distance, while the non-hierarchical distribution may not have. Therefore, if a set of non-hierarchically distributed samples becomes hierarchically distributed after an exponential map, the exponential map should have a strong distance expansion ability, which contradicts the content in \textbf{Remark 1}.\\
\indent Therefore, \textbf{Remark 1} implies that the hyperbolic discriminant objective makes the backbone extract the hierarchical information.

\subsection{Suffering from Deep Hyperbolic Representation Learning}
Firstly, we introduce the definition of the Lipschitz constant of the function $f(\textbf{I})$, which is presented as follows:
\begin{equation}
    Lip(f) = \sup_{\forall \textbf{I}_i,\textbf{I}_j}\frac{\Vert f(\textbf{I}_i)-f(\textbf{I}_j) \Vert_2}{\Vert \textbf{I}_i - \textbf{I}_j\Vert_2}.
\end{equation}
The Lipschitz constant of a backbone is widely used to explore the learning ability of the backbone. For example, in \cite{krishnan2020lipschitz,fazlyab2024certified}, they propose a regularization term to reduce the Lipschitz constant of the backbone, and increase the generalization ability of the backbone as a consequence. \cite{fazlyab2019efficient} employs the Lipschitz constant of the projection to estimate the testing error of the model. \cite{song2021adaptive} indicates if the objective function requires a very large Lipschitz constant, the model training will collapse. All those works suggest that we should avoid enlarging the Lipschitz constant of the backbone when we train the deep models.\\
\indent As discussed in the previous section, the hyperbolic discriminant objective directly forces the backbone to extract the hierarchy information, which also leads to a very large distance between samples. This means the Lipschitz constant of the backbone in hyperbolic representation learning is required to be large, which may harm the generalization ability and the training of the backbone.
\section{The Proposed Approach}
\subsection{Motivation}
\indent In this section, we show that bilinear pooling and its variants can non-linearly increase the distance between two input samples. In this way, if we let the hierarchy information be formed in the space of bilinear pooling, its inputted samples can avoid having large distances. Therefore, when the backbone in hyperbolic representation learning is followed by a bilinear pooling, the Lipschitz constant of the backbone can be reduced.\\
\indent Given a set of samples $\{\textbf{x}_i \in \mathbb{R}^{n\times 1}\}_{i=1}^L$, the bilinear pooling is calculated by an outer product, i.e., $\textbf{b}_i =vec(\textbf{x}_i\textbf{x}_i^T)\in \mathbb{R}^{n^2\times 1}$. Because $\textbf{b}_i^T\textbf{b}_j = (\vert \textbf{x}_i^T\textbf{x}_j\vert)^2$, the bilinear pooling can be seen as the explicit projection of the second-order polynomial kernel function $\kappa(\textbf{x}_i,\textbf{x}_j) = (\vert \textbf{x}_i^T\textbf{x}_j\vert + d)^2$ with $d = 0$. Imitating the polynomial kernel function, we can construct a general kernel function $\kappa(\textbf{x}_i,\textbf{x}_j) = f(\vert \textbf{x}_i^T\textbf{x}_j\vert )$ where $f(a) >0$ is a non-linearly increasing function. (The proof of $f(\vert \textbf{x}_i^T\textbf{x}_j\vert)$ is a kernel function is presented in the Appendix.) In this way, we analyze how bilinear pooling increases the distance of its inputted samples from the perspective of the general kernel function.
\begin{tcolorbox}
\textbf{Theorem 2.} If $f(a)>0$ is a non-linearly increasing function with respective to $a$, determining a kernel function $\kappa(\textbf{x}_i,\textbf{x}_j) = f(\textbf{x}_i^T\textbf{x}_j)$, there exists another increasing function $g(a)$ to let  $d_{\phi}(\textbf{x}_i,\textbf{x}_j) > g(\Vert \textbf{x}_i - \textbf{x}_j\Vert_2)$.
\end{tcolorbox}
The proof is presented in the Appendix.

\indent The above theorem demonstrates that the kernel function can non-linearly increase the distance between its inputted samples. As $f(a) = a^2 (a>0)$ is an increasing function, the second-order polynomial kernel function satisfies the requirement of the above theorem. Therefore, bilinear pooling can non-linear increase the distance of its input samples. Besides the second-order polynomial function, we can set $f(a) = \exp(ra) (r>0)$ to generate other kernel function that meets the requirement in Theorem 2. Therefore, Theorem 2 gives an instruction to construct other types of kernel functions that can increase the distances of their inputted samples, which extends the spectrum of the section of kernel functions.\\
\indent In deep learning, approaches prefer the explicit projection rather than the implicit one, like the general kernel function. This problem can be solved using the random Maclaurin projection, which approximates the product-based kernel function with a relatively low-dimensional feature compared with the dimensionality of the bilinear pooling \cite{gao2016compact}. However, \cite{yu2021fast} points out that the random Maclaurin projection is unstable because of the involvement of the random projection. To make the result stable, the dimension of the random Maclauri projection is relatively large.\\
\indent The calculation of random Maclaurin projection (RMP) has the same formulation as the low-rank Hadamard product-based bilinear projection (LK-HPBP) \cite{gao2016compact}, i.e.,
\begin{equation}
\label{equation_bili}
\textbf{b}_i = \textbf{W}_1^T\textbf{x}_i \circ \textbf{W}_2^T\textbf{x}_i
\end{equation}
In RMP, $\textbf{W}_1\in \mathbb{R}^{n\times h}$ and $\textbf{W}_2 \in \mathbb{R}^{n\times h}$ are two random projections, while they are learnable in the LK-HPBP. In the literature, the dimensionality of LK-HPBP is about 2048, while that of RMP is about 8194. This means that the learnable projections can output more stable and low-dimensional bilinear features.\\
\indent However, in practice, the experimental results of adopting the LK-HPBP into hyperbolic representation learning are not good. This is because it loses the ability of the distance expansion mentioned in Theorem 2, which makes the LK-HPBP less effective in capturing the hierarchy information. This is theoretically supported by the following theorem 3.
\begin{tcolorbox}
\textbf{Theorem 3.} If a set of bilinear features are linearly separated, their dimension can by reduced by a linear projection without harming the discriminant ability, and the distances among the obtained low-dimensional features are upper-bounded by a constant.
\end{tcolorbox}
\indent The proof is presented in the Appendix.\\
\indent In the following section, we want to introduce a novel approach to utilize the benefits of the two types of compact bilinear pooling and discard their shortcomings.
\subsection{Kernel approximation regularization}
\indent We also employ the Hadamard product bilinear projection like Eq.(\ref{equation_bili}) to learn the low-dimension bilinear features because we want the projection to have the solution able to expand the distance of its inputs. To improve the learning ability, we adopt a multi-head strategy to increase the number of learnable parameters. The projection is presented as follows:
\begin{equation}
\label{me_low_rank}
\textbf{b}_i = \textbf{P}^T\sum_{r=1}^K(\textbf{U}^T_r\textbf{x}_i \circ \textbf{V}^T_r\textbf{x}_i),
\end{equation}
where $\textbf{U}_r\times \mathbb{R}^{m\times h}$ and $\textbf{V}_r \in \mathbb{R}^{m\times h}$, and $\textbf{P}\in \mathbb{R}^{h\times l_1}$ are projection matrices, $K$ is the number of heads.\\
\indent To prevent Eq.(\ref{me_low_rank}) from losing the distance expansion power, we introduce the kernel approximation regularization denoted by ($\mathbb{KR}$), which is presented as follows:
\begin{equation}
\mathbb{KR}(\mathcal{I}) = \frac{1}{N_b^2}\sum_{i=1}^{N_b}\sum_{j=1}^{N_b} \vert \textbf{b}_i^T\textbf{b}_j - \kappa(\textbf{x}_i,\textbf{x}_j) \vert^2,
\end{equation}
where $\textbf{b}_i$ is calculated by Eq.(\ref{me_low_rank}), $N_b$ is the number of the samples in a mini-batch.
The implementation of the proposed approaches are depicted in Figure \ref{fig:architecuture}.\\\vspace{-5pt}\\
\textbf{Remark 2.} \emph{Our approach does not involve the normalization strategy, which is a must step of the RM projection. The reason consists of two aspects: (1) our bilinear features are learned by learnable Hadamard-product bilinear projection, which does not need the normalization strategy; (2) the role of the normalization strategy is to adjust the non-linearity of the bilinear pooling, which can be done in our method by adjusting the effect of kernel projection.}\\
%
%
\section{Experimentation}
%
\begin{table}[t]
\centering
\footnotesize
\setlength{\tabcolsep}{2.5pt}
 \renewcommand{\arraystretch}{0.7}
\caption{Description of adopted Datasets. $^\textbf{(*)}$The $\delta$ is approximated by subsampling k=50000 random nodes. 
}
\begin{tabular}{lccccc}
\toprule
\textsc{Dataset} & \textsc{$\delta$}  & \textsc{\#Nodes} & \textsc{\#Edges} & \textsc{\#Classes} & \textsc{\#Features}   \\ 
\midrule

Airport & 1.0 &  3,188 & 18,631 & 4 & 4 \\
Pubmed & 3.5  & 19,717 & 44,338 & 3  & 500 \\ 
Citeseer & 5  & 3,327 & 4,732 & 6 & 3,703 \\
Cora & 11.0  & 2,708 & 5,429 & 7 & 1,433  \\

\arrayrulecolor{black!30}\midrule
OGB Arxiv & 3.5$^{*}$ & 169,343  &  1,16M	 & 40 & 128 \\
OGB Products & 1.5$^{*}$ & 2,44M  &  61,85M & 47 & 100 \\


\arrayrulecolor{black}\bottomrule
\end{tabular}

\label{table_2}
\end{table}
We evaluate the effectiveness of our proposed hyperbolic bilinear projection and the regularization term on graph-structured data due to the richness of hierarchical features these datasets present. In fact, our algorithm can also be used in other domains such as image classification \cite{ermolov2022hyperbolic} or text summarization \cite{song2022preliminary}, but due to the page limit, we leave this analysis for future work.
\subsection{Experimental Setting}
\setlist[itemize]{itemindent=\dimexpr\labelwidth+\labelsep\relax, leftmargin=10pt}

\textbf{Datasets.} We adopt six datasets in the experiments. The flight network dataset, \textsc{Airport}~\cite{zhang2018link}; four citation datasets, \textsc{Cora}, \textsc{CiteSeer}, \textsc{Pubmed}~\cite{sen2008collective}, the OGB \textsc{Arxiv}; and a purchasing dataset, OGB \textsc{Products}~\cite{hu2020ogb}. The OGB datasets are notably larger, particularly regarding the number of classes. In summary, the citation datasets represent citation networks, where each node corresponds to a research paper, and edges capture the citation relationships. Each paper is associated with a categorical label indicating the topic of the publication. Similarly, in the flight network dataset, nodes represent airports, and edges denote operational traffic routes, with labels corresponding to activity levels. Lastly, the Products dataset represents product categories sold online, with edges between two products indicating that they are often purchased together. The statistics of these datasets are shown in Table~\ref{table_2}. As seen from the table, those data sets are diverse in terms of the number of classes, scale, and hyperbolicity. The hyperbolicity is measured as Gromov's $\delta > 0$, which indicates a clear hierarchy if the value is small, i.e., $\delta = 0$ represents a complete tree-like structure. \\
\indent \textbf{Baselines.} We compare our approach against a diverse set of baseline GNNs, encompassing both Euclidean models, such as GCN \cite{kipf2016semi}, GCNII \cite{chen2020simple}, GAT \cite{velickovic2017graph}, GraphSAGE \cite{hamilton2017inductive}, and Transformer~\cite{shi2020masked}. We also compare our method with hyperbolic versions of GNNs, such as HGCN\cite{chami2019hyperbolic} which modifies GCN layers using hyperbolic projections; H2H-GCN~\cite{dai2021hyperbolic}, that improves over HGCN operating directly in hyperbolic space; and deeper networks as DeepHGCN~\cite{liu2023deephgcn} that bring the concept of GCNII into the hyperbolic space. At last, we adopt two attentional variants, HAT~\cite{zhang2021hyperbolic} and GIL~\cite{zhu2020graph}.
\indent \textbf{Metrics}. We benchmark two alternatives, the hyperbolic bilinear pooling (HBP), and low-dimensional variant (HBP-$\mathbb{KR}$) on the following tasks: \underline{Node classification} which is evaluated by classification accuracy; \underline{Node Clustering}, for providing insights into the cohesion of clusters, which is measured using the Normalized Mutual Information (NMI); \underline{Link prediction} which is evaluated by AUROC metric following the protocols presented in ~\cite{chami2019hyperbolic}.
\indent \textbf{Training.} For a fair comparison, the learning rate, weight decay, dropout, activation functions, and number of layers/attention heads in the backbone are set as the same for all the methods. The details on the hyperparameter configuration are provided in Appendix.\\
\begin{table}[t]
\centering
\scriptsize
\setlength{\tabcolsep}{7pt}
\renewcommand{\arraystretch}{0.8}
\caption{Comparative results on node clustering reported as Normalized Mutual Information (NMI). Results are provided in the \textsc{Public split} for 5 random seeds.  `dim.' represents the dimension.}
    \begin{tabular}{lcccc}
        \toprule
        \textbf{Method} & \textbf{Airport} & \textbf{PubMed} & \textbf{CiteSeer} & \textbf{Cora}\\
        \midrule
        
        GCN & 47.0$\pm$0.5 & 39.3$\pm$0.6 & 44.5$\pm$0.6 & 60.3$\pm$1.2\\
        GCNII & 54.7$\pm$1.8 & 39.8$\pm$0.5 &  45.7$\pm$0.6 & {66.3$\pm$0.2} \\
        GAT & 51.6$\pm$0.3 & 37.3$\pm$1.2 & 44.3$\pm$0.7  & 58.1$\pm$1.1\\
        GraphSAGE & 50.9$\pm$0.6 & 35.7$\pm$1.3 & 43.8$\pm$0.5 & 59.9$\pm$0.7\\
        TransformerConv & 52.7$\pm$0.6 & 34.6$\pm$0.9 & 45.7$\pm$0.5 & 66.3$\pm$0.3\\
        
        \arrayrulecolor{black!30}\cmidrule{1-5}
        HGCN$^\textbf{(*)}$ & 33.7$\pm$1.1 & 32.7$\pm$0.9 & - & 57.6$\pm$0.7\\
        HAT & - & 39.3$\pm$0.7 & 43.9$\pm$0.6 & 58.2$\pm$1.0\\
        
        \arrayrulecolor{black!30}\cmidrule{1-5}
        HBP ($64^2$dim) & {70.0$\pm$0.7} & \textbf{41.3$\pm$0.5} & {46.8$\pm$0.3} & 65.5$\pm$0.3\\
        HBP-$\mathbb{KR}$ ($256$dim) & \textbf{73.6$\pm$0.5} & 41.1$\pm$0.5 & \textbf{47.4$\pm$0.1} &   {\textbf{66.6$\pm$0.4}}\\
        \arrayrulecolor{black}\bottomrule
    \end{tabular}
\small
$^\textbf{(*)}$Reproduced from the original work without using pretraining.
\label{table_8}
\end{table}
%
%
\begin{table*}[t]
\centering
\scriptsize
\setlength{\tabcolsep}{4pt}
\renewcommand{\arraystretch}{0.8}
\caption{\textsc{auroc} on the Link Prediction (\textsc{LP}) task and \textsc{accuracy}(\%) in the multi-class Node Classification (\textsc{NC}) tasks. Results are provided in the \textsc{Public split} for 5 random seeds.  `dim.' represents the dimension.}
    \begin{tabular}{clcccccccc}
        \toprule
        &\begin{tabular}{c}\textbf{Dataset} \\\textbf{Hyperbolicity}($\delta$)\end{tabular} & \multicolumn{2}{c}{\begin{tabular}{c}\textbf{Airport} \\($\delta = 1$)\end{tabular}} & \multicolumn{2}{c}{\begin{tabular}{c}\textbf{PubMed} \\($\delta = 3.5$)\end{tabular}}  & \multicolumn{2}{c}{\begin{tabular}{c}\textbf{CiteSeer} \\($\delta = 5$)\end{tabular}} & \multicolumn{2}{c}{\begin{tabular}{c}\textbf{Cora} \\($\delta =11$)\end{tabular}}\\
        \cmidrule(lr){3-4} \cmidrule(lr){5-6} \cmidrule(lr){7-8} \cmidrule(lr){9-10}
        &\textbf{Task} & \textbf{LP} & \textbf{NC} & \textbf{LP} & \textbf{NC} & \textbf{LP} & \textbf{NC} & \textbf{LP} & \textbf{NC} \\
        \midrule
        
        \parbox[t]{3mm}{\multirow{5}{*}{\rotatebox[origin=c]{90}{\textsc{Euclidean}}}} &GCN             & 97.1$\pm$0.4 & 81.5$\pm$0.6 & 89.5$\pm$3.6 & 79.3$\pm$0.3 & 82.5$\pm$1.9 &  71.3$\pm$0.7 & 90.4$\pm$0.2 & 81.4$\pm$0.6 \\
        &GCNII           & 96.2$\pm$0.5 & 83.7$\pm$0.9  & 96.5$\pm$0.3  &  79.5$\pm$0.4 & 92.3$\pm$0.5  & 71.6$\pm$0.4 & 96.1$\pm$0.5 &  84.4$\pm$0.3\\
        &GAT             & 95.8$\pm$0.2 & 82.7$\pm$0.4 & 91.4$\pm$1.8 & 78.2$\pm$0.5 & 86.5$\pm$1.5 & 70.9$\pm$0.8 & 93.2$\pm$0.2 & 80.2$\pm$0.5 \\
        &GraphSAGE       & 93.4$\pm$0.5 & 82.2$\pm$0.4 & 86.2$\pm$0.8 & 77.2$\pm$0.7 & 92.0$\pm$0.4 & 71.2$\pm$0.4 & 85.5$\pm$0.5 & 81.2$\pm$0.5 \\
        &TransformerConv & 96.4$\pm$0.4 &  83.3$\pm$0.1 & 84.6$\pm$1.1  & 76.6$\pm$0.5  & 78.2$\pm$1.4  & 70.7$\pm$0.4  & 87.9$\pm$0.9 &  80.1$\pm$0.5\\
        \arrayrulecolor{black!30}\cmidrule{1-10}
        
        \parbox[t]{3mm}{\multirow{5}{*}{\rotatebox[origin=c]{90}{\textsc{Hyperbolic}}}}&HGCN~\cite{chami2019hyperbolic} & 96.4$\pm$0.1 & 90.6$\pm$0.2 & 96.3$\pm$0.1 & 80.3$\pm$0.3 & 96.6$\pm$0.0 & 68.0$\pm$0.6 & 92.9$\pm$0.1 & 79.9$\pm$0.2 \\
        &HAT~\cite{zhu2020graph} & 97.9$\pm$0.1 & 89.6$\pm$1.0 & 94.2$\pm$0.2 & 77.4$\pm$0.7 & 95.8$\pm$0.4 & 68.6$\pm$0.3 & 94.0$\pm$0.2 & 78.3$\pm$1.4 \\
        &GIL~\cite{zhu2020graph} & {98.8$\pm$0.3 }& 91.3$\pm$0.8 & 95.5$\pm$0.2 & 78.9$\pm$0.3 & \textbf{99.8$\pm$0.4 }& 73.0$\pm$0.7 & {98.3$\pm$0.8} & 83.6$\pm$0.6 \\
        &H2H-GCN~\cite{dai2021hyperbolic} & 96.7$\pm$0.1 & 91.0$\pm$0.3&97.1$\pm$0.1 & {82.3$\pm$0.4} & - & - &95.4$\pm$0.1& 83.6$\pm$0.8\\
        &{DeepHGCN}~\cite{liu2023deephgcn} & {98.1$\pm$0.3} & {94.7$\pm$0.9} & {96.1$\pm$0.2} & {79.4$\pm$0.9} & {97.4$\pm$0.4} & {73.3$\pm$0.7} & {93.9$\pm$0.8} & {83.6$\pm$0.4} \\
        \arrayrulecolor{black!30}\cmidrule{1-10}
        
        \parbox[t]{3mm}{\multirow{2}{*}{\rotatebox[origin=c]{90}{\textsc{Ours}}}}&HBP ($64^2$dim.) & 97.4$\pm$0.4& 91.8$\pm$0.1& {97.7$\pm$0.2} & 80.3$\pm$0.6 & 99.1$\pm$0.3 & {73.3$\pm$0.3} & {\textbf{98.4$\pm$0.4}} & {84.8$\pm$0.4}\\
        &HBP-$\mathbb{KR}$ ($256$dim.) & \textbf{98.9$\pm$0.1} & \textbf{94.9$\pm$0.2} & \textbf{97.8$\pm$0.2} & \textbf{82.7$\pm$0.6} & {99.2$\pm$0.3} & \textbf{74.4$\pm$0.4} & 98.3$\pm$0.1  &   \textbf{85.6$\pm$0.3} \\
        \arrayrulecolor{black}\bottomrule
    \end{tabular}
\label{table_7}
\end{table*}
\begin{table}[t]
\centering
\scriptsize
\setlength{\tabcolsep}{8pt}
\renewcommand{\arraystretch}{0.8}
\caption{\textsc{Accuracy}(\%) in the multi-class node classification task on the OGB. The baseline corresponds to GraphSAGE in OGB ArXiv and OGB GAT in the Products dataset. `dim.' represents the dimension.}
\begin{tabular}{lcc} 
\arrayrulecolor{black}\toprule
{\textbf{Method}} & \textbf{OGB ArXiv} &\textbf{OGB Products} \\

\arrayrulecolor{black}\midrule

 Baseline ($128$dim.) & 69.1$\pm$0.1 & 77.1$\pm$0.6 \\
\arrayrulecolor{black!30}\cmidrule{1-3}
 HBP ($128^2$dim.) & 72.3$\pm$0.2 & 78.9$\pm$0.2\\
  HBP-$\mathbb{KR}_\text{LOW-RANK}$ ($512$dim.)  & \textbf{73.4$\pm$0.2}  & \textbf{79.8$\pm$0.3}\\

\arrayrulecolor{black}\bottomrule
\end{tabular}

\label{table_12}
\end{table}

\subsection{Comparison with State-of-the-art} 
We set the batch size as $64$ for all datasets. The coefficient of the regularization term is tuned in the grid $\{10^{-4},10^{-3},10^{-2},10^{-1}\}$ in the experiments. All the results are based on the \textsc{Public split} from~\cite{yang2016revisiting}. We select GCNII~\cite{chen2020simple} as the backbone of our method. For the small datasets, we summarize the main results of the classification and clustering tasks for the citation datasets in Tables~\ref{table_7} and~\ref{table_8}, respectively. \\
\indent As seen from Table~\ref{table_8}, we can find our approach achieves the best performance on the clustering task. Especially, the bilinear pooling can increase GCNII's performance to $18.9\%$ on the airport data set. Another interesting observation is that HGNN is much worse than the GNN in the clustering task, which may indicate that the training of hyperbolic version models is difficult as it has too many hyperparameter to tune, such as the root point and curvatures in different layers. Considering our approach consistently surpasses the GCNII, which may be because it only has one hyperbolic layer. Thus, the training of our model is much easier.\\
\indent As seen in Tables~\ref{table_7}, our method still achieves the best performance on most datasets, compared with the hyperbolic methods. This validates the effectiveness of the proposed algorithm. If we compare the BP with the rest method, it still achieves good results on some datasets. This validates that bilinear pooling can help hyperbolic representation learning. Our low-dimensional version can surpass the naive bilinear pooling, which may be because our proposed regularization can adjust the non-linearity of the bilinear pooling, while the naive one can not because it does not adopt the normalization strategy. This demonstrates the superiority of the proposed method.\\
\indent For the OGB datasets (large datasets), the results are summarized in Table~\ref{table_12}. Because there is less literature reporting the performance of hyperbolic methods on the two datasets, we run the public code to obtain the results by ourselves. However, it may be because those deep hyperbolic structures are hard to tune, we do not successfully obtain good results. For the non-hyperbolic methods, we only present the deep structure with the best results. Specifically, the baseline used for the OGB ArXiv dataset is GraphSAGE, while GAT is used for the OGB-Products dataset. As a comparison, we implement our approach to the two structures with $512$ dimension. We can see that our method still outperforms comparison methods. More details are presented in the appendix.

\begin{figure*}[t]
    \centering
    \footnotesize
        
    \includegraphics[width=0.8\textwidth]{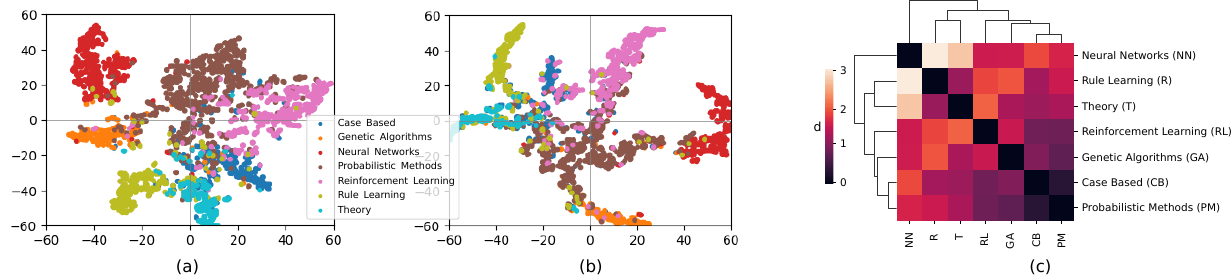}
    \caption{t-SNE visualization of the embeddings for (a) the Euclidean model and (b) the model incorporating the hyperbolic projection in the Cora dataset. The hierarchical structure of the dataset becomes visible after applying the hyperbolic projection. In (c), the embeddings are hierarchically clustered according to the class centroid distances obtained from (b).}
    \label{figure_4}
\end{figure*}

\begin{figure}[t]
    \centering
    \footnotesize
    \includegraphics[width=0.24\textwidth]{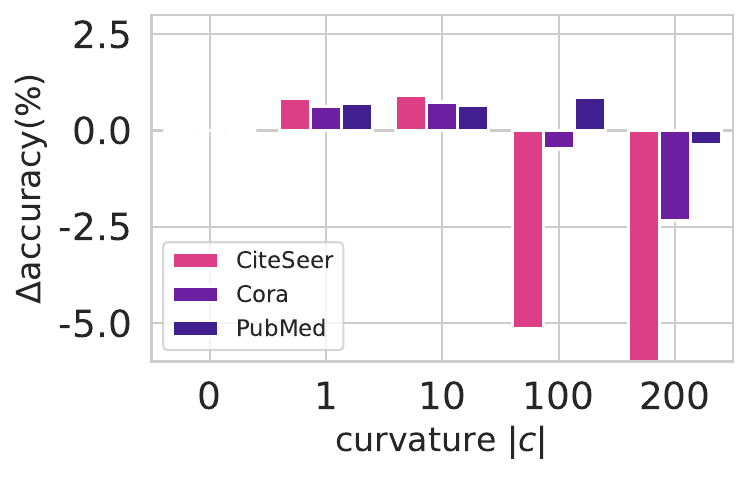}
\caption{Impact of the curvature upon different levels of curvature in the accuracy for the citation datasets.}
    \label{figure_2}
\end{figure}

\subsection{Ablation Experiments}

\textbf{Visualization of the embedding.} We present t-SNE visualizations of the embeddings produced by the backbone in Figure~\ref{figure_4}(a) for the Euclidean model, (b) for the model incorporating hyperbolic projection, and (c) shows the hierarchy in (b) using the Nearest Point algorithm. As seen from the Figures, the Euclidean model primarily discriminates classes in the Cora dataset by angle, failing to capture the hierarchy effectively. In contrast, using the hyperbolic discriminant objective, the hierarchical structure within the extracted features becomes apparent. This validates our analysis in Section 2.2.

\textbf{Impact of the curvature.} In our analysis, the bound for hyperbolic distance is close to the curvature. Thus, we present results of different curvatures of the hyperbolic space in Figure~\ref{figure_2} with the GCNII architecture of our proposed method on the citation datasets. As seen from the Figure, we can observe a consistent improvement (approximately by 1\%) from the reference accuracy in model performance by setting $c\in \{1,10\}$. This means that the performance is not sensitive to the curvature. This may be because the hierarchical information is captured by the backbone rather than the hyperbolic projections. When the curvature $c\in \{100,200\}$, the performance is dropped significantly. This may be because the large curvature makes the computation of the hyperbolic distance hard. Therefore, we set the curvature $1$ in the experiments.

\textbf{Hyperparameters on the Kernel regularization:} We set the parameter of the kernel regularization as $0.1$. Since BP quadratically expands the size of the features extracted by the backbone, learning a compact bilinear representation is key to applications. We explore different feature dimensions for the kernel approximation. We evaluate dimensions \( h = [128, 256, 512] \), results are shown in Table~\ref{table_13}. We find that as the dimension increases, the performance of the method reaches a point of saturation at $256$, which is much smaller than $64^2$. This indicates that our learning framework effectively reduces the dimension of bilinear features. We also examine different kernel functions in the kernel regularization. We find that the polynomial kernel has the best performance.

\begin{table}[t]
\centering
\scriptsize
\setlength{\tabcolsep}{5pt}
\renewcommand{\arraystretch}{0.8}
\caption{Ablation experiments on hyperparameter tuning on the proposed Kernel regularization.}
\begin{tabular}{lccc} \toprule

   &\textbf{Cora}  & \textbf{CiteSeer} &  \textbf{PubMed} \\ 
\midrule
\rowcolor{black!15}  \textsc{Kernel Regulatization}     &   &   &    \\
 HBP-$\mathbb{KR}$ ($128$dim)            & 84.8$\pm$0.5 & 71.7$\pm$0.7  &  79.7$\pm$0.5   \\
 HBP-$\mathbb{KR}$ ($256$dim)            &  \textbf{85.6$\pm$0.3} & \textbf{74.4$\pm$0.4}  &  \textbf{79.7$\pm$0.6} \\
 HBP-$\mathbb{KR}$ ($512$dim)            &   85.3$\pm$0.3 & 74.3$\pm$0.4  &  79.6$\pm$0.5    \\
  \rowcolor{black!15}  \textsc{Kernel Functions}     &   &   &    \\
  HBP-$\mathbb{KR}_{256dim}$ [$f = (\textbf{x}_i^T\textbf{x}_j)^2$]    &  85.6$\pm$0.3 & 74.4$\pm$0.4  &  79.7$\pm$0.6  \\
  HBP-$\mathbb{KR}_{256dim}$ [$f = \exp(\vert \textbf{x}_i^T\textbf{x}_j)\vert)$]                           & 84.9$\pm$0.4 & 71.8$\pm$0.6  &   77.3$\pm$1.3\\

\bottomrule
\end{tabular}
\label{table_13}
\end{table}
%

\begin{table}[t]
\centering
\scriptsize
\setlength{\tabcolsep}{4pt}
\renewcommand{\arraystretch}{0.8}
\caption{Node classification accuracy ($\%$) comparison of hyperbolic bilinear pooling applied on different graph architectures. In addition to the previously reported GCNII convolution layer, we include the classical GCN and Transformer convolution.}
\begin{tabular}{llccc} 
\arrayrulecolor{black}\toprule
{\textsc{Full-Split}} &
{\textsc{Method}} & \textsc{GCN} &\textsc{GCNII-Conv}   & \textsc{Transf-Conv} \\

\arrayrulecolor{black}\midrule
\multirow{3}{*}{{\textbf{Cora}}}
&Euclidean  & 86.3$\pm$0.5 & 82.1$\pm$3.1 & 85.6$\pm$0.4 \\
&BP         & 84.3$\pm$0.4 & 85.3$\pm$0.9 & 83.8$\pm$1.4 \\
&HBP        & 86.9$\pm$0.2 & 87.7$\pm$0.3 & 86.1$\pm$0.3 \\
&HBP-$\mathbb{KR}$      & \textbf{87.1}$\pm$0.4 & \textbf{88.3$\pm$0.5} & \textbf{86.7$\pm$0.2} \\
\arrayrulecolor{black!30}\cmidrule{1-5}
\multirow{3}{*}{{\textbf{Cireseer}}}
&Euclidean  & 78.3$\pm$0.2 & 78.1$\pm$0.2 & 77.4$\pm$1.4 \\
&BP         & 77.4$\pm$0.1 & 76.2$\pm$0.6 & 72.8$\pm$1.1 \\
&HBP        & 78.8$\pm$0.1 & 78.8$\pm$1.0 & 78.0$\pm$0.5 \\
&HBP-$\mathbb{KR}$      & 78.9$\pm$0.5 & \textbf{79.2$\pm$0.4} & \textbf{79.8$\pm$0.5} \\
\arrayrulecolor{black!30}\cmidrule{1-5}
\multirow{3}{*}{{\textbf{Pubmed}}}
&Euclidean  & 88.0$\pm$0.2 & 86.7$\pm$0.3 & 87.9$\pm$0.2 \\
&BP         & 86.6$\pm$0.4 & 87.5$\pm$0.2 & 87.9$\pm$0.3 \\
&HBP        & 88.3$\pm$0.2 & 87.4$\pm$0.2 & 89.1$\pm$0.2 \\
&HBP-$\mathbb{KR}$     & \textbf{88.9$\pm$0.6} & \textbf{88.2$\pm$0.1} & \textbf{89.1$\pm$0.2} \\
\arrayrulecolor{black}\bottomrule
\end{tabular}

\label{table_9}
\end{table}

\textbf{Comparison among different GNN architectures:}
We conducted the evaluation of hyperbolic bilinear pooling across different GNN backbones, including GCN, GCNII, and Transformer architectures. We test different bilinear pooling techniques, such as bilinear pooling without hyperbolic analysis (BP), bilinear pooling with hyperbolic analysis (HBP), and low-dimensional bilinear pooling with hyperbolic analysis (HBP-$\mathbb{KR}$). The summarized results are available in Table~\ref{table_9}. As seen from the results, we can find that our proposed method can achieve the best results on most of the datasets, which demonstrates the flexibility of the proposed method for different backbones.\\
\indent Besides, we also observe that the performance of bilinear pooling (BP) is bad in most of the datasets and backbones, even worse than the Euclidean GNN, not presenting any enhancement. However, when the BP is coupled with hyperbolic projection, the performance is significantly improved. The reason may be because the Euclidean representation separates different classes using angles. The distance expansion of bilinear pooling may harm the generalization ability of the feature, as it increases the intra-class variances. This will not happen in traditional bilinear pooling as it uses the $L_2$-normalization to project the bilinear features on a sphere, which only keeps the bilinear information in the perspective of the angles. This observation validates our analysis of why we adopt the bilinear pooling hyperbolic representation learning.
%
%
\section{Conclusions}
Our study introduces a novel approach to enhancing hyperbolic representation learning by incorporating bilinear pooling and hyperbolic discriminant analysis. Firstly, we reveal that hyperbolic representation learning induces the backbone to have a large Lipscthitz constant when capturing hierarchical information, harming the generalization ability of the model. To solve this problem, we demonstrate that bilinear pooling can enhance the distance of its input features, preventing the backbone's Lipschitz constant from being large. At last, we conduct extensive experiments to evaluate the proposed algorithm. The results validate the superiority of the proposed method and the mathematical analysis.

\bibliography{refs}

\newpage

\appendix

\textbf{{\LARGE Supplementary material: Enhance Hyperbolic Representation Learning via Second-order Pooling}}
\bigbreak


\section{Inverse bilinear pooling}
In this section, we explain how to calculate the distribution in the explanatory example in Figure~\ref{figure_inverser_bilinear_pooling} in the main manuscript of the paper. 

Given an example whose feature map is made of two two-dimensional vectors $\textbf{X} = [\textbf{x}_1,\textbf{x}_2] \in \mathbb{R}^{2 \times 2}$, the feature vector of this sample can be calculated by the average pooling $\textbf{x} = (\textbf{x}_1 + \textbf{x}_2)/2$ which is often adopted in the deep learning. Thus, the bilinear pooling on $\textbf{X}$ is calculated as,
\begin{equation}
    \textbf{Z} = \textbf{X
    }\textbf{X}^T  \in \mathbb{R}^{2 \times 2}.
\end{equation}

For easy demonstration, we suppose all the bilinear features are on a two-dimensional subspace of $\mathbb{R}^{2\times 2}$. This can be achieved by setting $\textbf{Z}_{22} = c$,  because $\textbf{Z}_{12} = \textbf{Z}_{21}$. This setting allows us to calculate the inverse of bilinear pooling.

Given a two-dimensional bilinear feature ${\textbf{z}} = [z_1,z_2] \in \mathbb{R}^{2}$, we can recover it to the matrix form $\textbf{Z}$ by using the equation:
\begin{equation}
    \textbf{Z} = \begin{bmatrix}
z_1 & z_2 \\
z_2 & c
\end{bmatrix},
\end{equation}
where $c>0$ is the fixed value.

Then, we decompose $ \textbf{Z} = \textbf{X}\textbf{X}^T$ via the singular value decomposition. The matrix $\textbf{X}$ corresponds to the inverse of bilinear pooling. In this figure, we set $c = 1$. We show the distribution of $\textbf{X}$ by the average pooling of $\textbf{X}$, which is also a $2$-dimensional vector.

\section{Proof of Theorem 1}
\begin{figure}[t]
    \centering
    \footnotesize
    \includegraphics[width=0.9\linewidth]{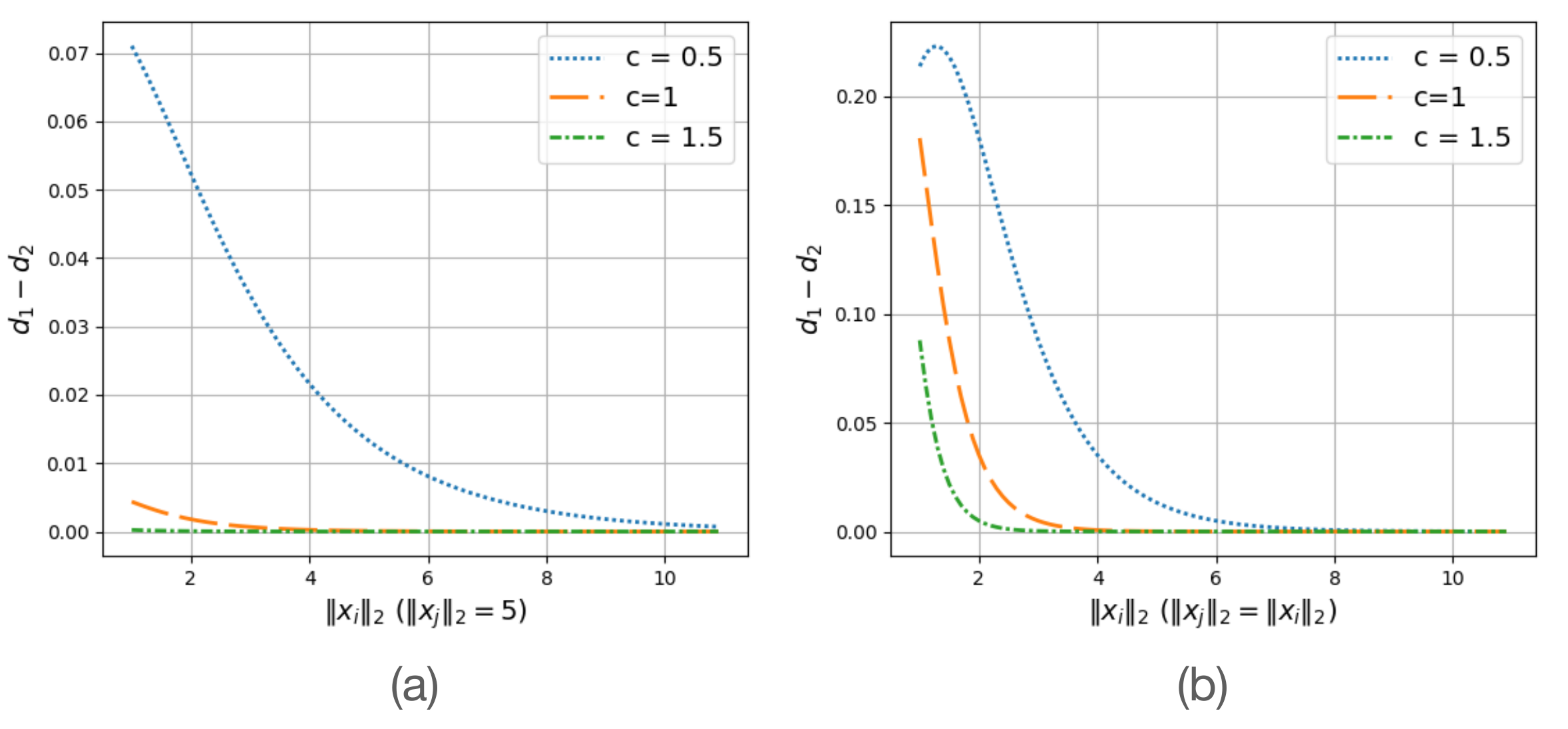}
\caption{\footnotesize{The error is calculated by $\tanh^{-1}(\sqrt{\frac{k_i^2+k_j^2}{1+k_i^2k_j^2}}) - \tanh^{-1}(\frac{k_i^2+k_j^2}{1+k_i^2k_j^2})$ with different $c$, where $k_i = \tanh(\sqrt{c}\Vert \textbf{x}_i\Vert_2/2)$}.}
    \label{figure_2_error}
\end{figure}

\begin{tcolorbox}
\textbf{Theorem 1.} Given two samples $\textbf{x}_i$ and $\textbf{x}_j$ in the tangent space of $\textbf{0}$, Eq.(\ref{exp_fun1}) projects them into the hyperbolic space denoted by $\textbf{z}_i$ and $\textbf{z}_j$, respectively, whose hyperbolic distance is $d_{c}^h(\textbf{z}_i,\textbf{z}_j)$. The relation with the distances satisfy:
\begin{equation}
\begin{split}
&d_{c}^h(\textbf{z}_i,\textbf{z}_j) \approx  \Vert \textbf{x}_i\Vert_2 + \Vert \textbf{x}_j \Vert_2, \quad \textbf{x}_i^T\textbf{x}_j = 0\\
&d_{c}^h(\textbf{z}_i,\textbf{z}_j) = \Vert \textbf{x}_i - \textbf{x}_j\Vert_2, \!\quad\! \textbf{x}_i = k \textbf{x}_j, k \in \mathbb{R}\\
\end{split}
\end{equation}
\end{tcolorbox}
\textbf{Proof}. The proof consists of 4 steps.
\\\vspace{-5pt}\\
\underline{\textbf{S}tep 1}:\\
\indent Before proving the theorem, we should introduce an important equation frequently used in the following proof.\\
\begin{equation}
\begin{split}
    &\tanh^{-1}(a) + \tanh^{-1}(b)\\
   =&\tanh^{-1}(\frac{a+b}{1+ab}) \quad a,b\in (-1,1)
    \end{split}
\end{equation}
Because $\tanh^{-1}(a) = \log(\frac{1+a}{1-a})$, this equation can be easily validated by the following procedure:
\begin{equation}
\begin{split}
    &\tanh^{-1}(a) + \tanh^{-1}(b) =\\
=& \log(\frac{1+a}{1-a}) + \log(\frac{1+b}{1-b})\\
=&\log(\frac{1+a}{1-a}\frac{1+b}{1-b})\\
=&\log(\frac{1+ab + (a+b)}{1+ab - (a+b)})\\
=&\log(\frac{1+(a+b)/(1+ab)}{1-(a+b)/(1+ab)})\\
=&\tanh^{-1}(\frac{a+b}{1+ab})
\end{split}
\end{equation}
\\\vspace{-5pt}\\
\underline{\textbf{S}tep 2:}\\
\indent For simplicity, we introduce some useful symbols in this step.\\
\indent Given two samples $\textbf{x}_i$ and $\textbf{x}_j$ in the tangent space of $\textbf{z}=0$, their projections in the hyperbolic space are $\textbf{z}_i = \exp_0(\textbf{x}_i)$ and $\textbf{z}_j = \exp_0(\textbf{x}_j)$. Because $\exp_0(\textbf{x}_i) = \frac{\tanh(\sqrt{c}{\Vert \textbf{x}_i\Vert_2}/{2})}{\sqrt{c}}\frac{\textbf{x}_i}{\Vert \textbf{x}_i\Vert_2}$, we can rewrite it for convenience as the following formulation:
\begin{equation}
    \textbf{z}_i = \frac{k_i}{\sqrt{c}} \textbf{e}_i
\end{equation} 
where $k_i = \tanh(\sqrt{c}\frac{\Vert \textbf{x}_i\Vert_2}{2})$ and $\textbf{e}_i =\frac{\textbf{x}_i}{\Vert \textbf{x}_i\Vert_2}$ is the direction of $\textbf{z}_i$ because $\Vert \textbf{e}_i \Vert_2 = 1$.\\
\indent For easy reading, we also present the definition of the hyperbolic distance between $\textbf{z}_i$ and $\textbf{z}_j$ here:\\
\begin{equation}
    d_c^h(\textbf{z}_i,\textbf{z}_j) = \frac{2}{\sqrt{c}}\tanh^{-1}(\sqrt{c}\Vert-\textbf{z}_i \oplus_c \textbf{z}_j\Vert_2)
\end{equation}
where 
\begin{equation}
    \textbf{z}_i \oplus \textbf{z}_j = \frac{(1 + 2c<\textbf{z}_i,\textbf{z}_j> + c\Vert \textbf{z}_j\Vert_2^2)\textbf{z}_i + (1-c\Vert \textbf{z}_i\Vert_2^2)\textbf{z}_j}{ 1 + 2c<\textbf{z}_i,\textbf{z}_j> + c^2\Vert\textbf{z}_i\Vert_2^2 \Vert \textbf{z}_j\Vert_2^2}
\end{equation}
\\\vspace{-5pt}\\
\indent \underline{\textbf{S}tep 3:}\\
\indent We prove the case $\textbf{z}_i^T\textbf{z}_j = 0$.\\
\indent We calculate the squared norm $\Vert -\textbf{z}_i \oplus_c \textbf{z}_j\Vert_2^2$ as follows:
\begin{equation}
\begin{split}
    & \Vert -\textbf{z}_i \oplus_c \textbf{z}_j\Vert_2^2 =\\
    =&\Vert \frac{-(1+c(k_j/\sqrt{c})^2)k_i\textbf{e}_i/\sqrt{c} + (1-c (k_i\sqrt{c})^2)k_j\textbf{e}_j/\sqrt{c}}{1 + c^2(k_i/\sqrt{c})^2(k_j/\sqrt{c})^2} \Vert_2^2\\
    =&\Vert \frac{-(1+(k_j)^2)k_i\textbf{e}_i + (1-(k_i)^2)k_j\textbf{e}_j}{(1 + (k_i)^2(k_j)^2)\sqrt{c}}\Vert_2^2
    \end{split}
\end{equation}
\indent Because $\textbf{x}_i\textbf{x}_j = 0 \Leftrightarrow \textbf{e}_i^T\textbf{e}_j = 0$, we have the squared norm
\begin{equation}
\begin{split}
    & \Vert -\textbf{z}_i \oplus_c \textbf{z}_j\Vert_2^2 =\\
    =&\Vert \frac{-(1+(k_j)^2)k_i\textbf{e}_i + (1-(k_i)^2)k_j\textbf{e}_j}{(1 + (k_i)^2(k_j)^2)\sqrt{c}}\Vert_2^2 \\
    =&{\frac{(1+(k_j)^2)^2k_i^2 + (1-(k_i)^2)^2k_j^2}{(1 + (k_i)^2(k_j)^2)^2{c}}}\\
    =& {\frac{k_i^2 + k_j^2}{(1+k_i^2k_j^2)c}}
    \end{split}
\end{equation}
 According to the above equation, because $k_i <1$ and $k_j < 1$, there are ${c}\Vert -\textbf{z}_i \oplus_c \textbf{z}_j\Vert_2^2<1$. Thus, $k_i^2<k_i$, $k_j^2<k_j$, and ${c}\Vert -\textbf{z}_i \oplus_c \textbf{z}_j\Vert_2^2 < \sqrt{c}\Vert -\textbf{z}_i \oplus_c \textbf{z}_j\Vert_2$. Because $\tanh^{-1}(a) + \tanh^{-1}(b) = \tanh^{-1}(\frac{a+b}{1+ab})$, we have $\tanh^{-1}(c\Vert -\textbf{z}_i \oplus_c \textbf{z}_j\Vert_2^2) = \tanh^{-1}(k_i^2) + \tanh^{-1}(k_j^2)$. Considering $y=\tanh^{-1}(x) (-1<x<1)$ is an increasing function, we have
\begin{equation}
\label{equation_appro_long}
\begin{split}
&\frac{\sqrt{c}}{2}d_c^h(\textbf{z}_i,\textbf{z}_j)\\
=&\tanh^{-1}(\sqrt{c}\Vert -\textbf{z}_i \oplus_c \textbf{z}_j\Vert_2)\\
>&\tanh^{-1}(c\Vert -\textbf{z}_i \oplus_c \textbf{z}_j\Vert_2^2)\\
=& \tanh^{-1}(k_i^2) + \tanh^{-1}(k_j^2)\\
<& \tanh^{-1}(k_i) + \tanh^{-1}(k_j)\\
=& \sqrt{c}\frac{\Vert \textbf{x}_i\Vert_2}{2} + \sqrt{c}\frac{\Vert \textbf{x}_j\Vert_2}{2}
\end{split}
\end{equation}
\\
\\
We want to replace the inequity symbols (i.e., $>$ and $<$) in the above equation with the approximation symbol ($\approx$). Therefore, we introduce approximations presented in the following equation if $k_i \approx 1$, $k_j \approx 1$:
\begin{equation}
\label{equation_approx}
\begin{split}
    \tanh^{-1}(\sqrt{\frac{k_i^2 + k_j^2}{1+k_i^2k_j^2}}) &\approx \tanh^{-1}(\frac{k_i^2 + k_j^2}{1+k_i^2k_j^2})\\
    \tanh^{-1}(k_i)& \approx \tanh^{-1}(k_i^2)\\
    \tanh^{-1}(k_j)& \approx \tanh^{-1}(k_j^2)
\end{split}
\end{equation}
\indent According to the definition of $k_i = \tanh(\sqrt{c}\Vert \textbf{x}_i\Vert_2/2)$, we know that $k_i < 1$. We list a set of value of $y = \tanh(x)$ as follows:
\begin{equation}
\label{equation_list_1}
    \begin{matrix}
        \tanh(3) = 0.99505475 & \tanh(3)^2 = 0.99013396\\
        \tanh(4) = 0.99932929 & \tanh(4)^2 = 0.99865904\\
        \tanh(5) = 0.99990920 &\tanh(5)^2 =  0.99981841\\
        \tanh(6) = 0.99998771 &\tanh(6)^2 =  0.99997542\\
    \end{matrix}
\end{equation}
The above values indicate that the speed of $k_i$ and $k_i^2$ approximating $1$ is very fast if $\sqrt{c}\Vert \textbf{x}_i\Vert_2/2>5$. Besides, we can find that $k_i$ also approaches $k_i^2$ very fast. As $k_i^2 \approx 1$ and $k_j^2 \approx 1$, we know that $\frac{k_i^2+k_j^2}{1+k_i^2k_j^2}$ also approximate 1 very fast.\\
\indent As the list in the Eq. (\ref{equation_list_1}), we know the approximation in the equation (\ref{equation_approx}) can be hold if $(\sqrt{c}\Vert \textbf{x}_i\Vert_2)/2 > 5$. We have also graphically presented the value of $\tanh^{-1}(\sqrt{\frac{k_i^2 + k_j^2}{1+k_i^2k_j^2}}) - \tanh^{-1}(\frac{k_i^2 + k_j^2}{1+k_i^2k_j^2})$ in Figure \ref{figure_2_error}.\\
\indent In deep learning, as the dimension of the backbone's outputs is often larger than $64$, therefore, the condition $(\sqrt{c}\Vert \textbf{x}_i\Vert_2)/2 > 5$ is easy to satisfy. This is because for a dense vector, its norm increases with the dimension.\\
\indent In this way, by replacing the inequality symbols ($>$ and $<$) in Eq.(\ref{equation_appro_long}) with the approximation symbol ($\approx$), we can have that
\begin{equation}
    d_c^h(\textbf{z}_j,\textbf{z}_i) \approx \Vert \textbf{x}_i\Vert_2 + \Vert \textbf{x}_j \Vert_2
\end{equation}
\\
\vspace{-5pt}
\\
\underline{\textbf{S}tep 4}:\\
\indent At last, we prove the second case $\textbf{x}_i = k\textbf{x}_j$.\\
\indent We also set $\textbf{z}_i = k_i \textbf{e}_i$ and $\textbf{z}_j = k_j \textbf{e}_j$. Because $\textbf{x}_i = k\textbf{x}_j$, there is $\textbf{e}_i = \textbf{e}_j$ if $k>0$; $\textbf{e}_i = -1\textbf{e}_j$ if $k<0$. Therefore, we let $\textbf{e}_i = \textbf{e}$ and $\textbf{e}_j = \bar{k} \textbf{e}$ where
\begin{equation}
    \bar{k} = \left\{
    \begin{matrix}
       & 1, k>0\\
       & -1, k<0
    \end{matrix}
    \right.
\end{equation}
By using the above defined symbols, we have the following equation:
\begin{equation}
\begin{split}
   & -\textbf{z}_i \oplus_c \textbf{z}_j\\
   =& \frac{(1-k_i^2)k_j\bar{k}/\sqrt{c}- (1-2k_ik_j\bar{k} + \bar{k^2}k_j^2)k_i/\sqrt{c}}{1-2k_ik_j\bar{k} + k_i^2k_j^2\bar{k}^2}\textbf{e}\\
   =&\frac{(k_j\bar{k}-k_i)(1-k_ik_j\bar{k})}{(1-k_ik_j\bar{k})^2\sqrt{c}}\textbf{e}\\
   =& \frac{k_j\bar{k} - k_i}{(1-k_ik_j\bar{k})/\sqrt{c}}\textbf{e}
\end{split}
\end{equation}
\indent Because $\Vert \textbf{e} \Vert_2 = 1$, the hyperbolic distance is:
\begin{equation}
\begin{split}
    &d_c^h(\textbf{z}_i,\textbf{z}_j) =  \frac{2}{\sqrt{c}}\tanh^{-1}(\sqrt{c}\Vert -\textbf{z}_i \oplus_c \textbf{z}_j \Vert_2)\\
    &= \frac{2}{\sqrt{c}}\tanh^{-1}(\vert \frac{k_j\bar{k} - k_i}{1-k_ik_j\bar{k}}\vert)\\
\end{split}
\end{equation}
\indent In the following content, we also use the equation $\tanh^{-1}(\frac{a+b}{1+ab}) = \tanh^{-1}(a) + \tanh^{-1}(b)$ to prove the conclusion.\\ 
\indent Without losing generality, we suppose $k_j > k_i$. Thus, if $\bar{k} = 1$, the distance is $ d_c^h(\textbf{z}_i,\textbf{z}_j) = \frac{2}{\sqrt{c}}\tanh^{-1}( \frac{k_j - k_i}{1-k_ik_j}) = \frac{2}{\sqrt{c}}(\tanh^{-1}(k_j) + \tanh^{-1}(-k_i))$. Because $\tanh^{-1}(-a) = \tanh^{-1}(a)$, there is $d_c^h(\textbf{z}_i,\textbf{z}_j) = \frac{2}{\sqrt{c}}(\tanh^{-1}(k_j) - \tanh^{-1}(k_i)) = \frac{2}{\sqrt{c}}(\sqrt{c}/2)(\Vert \textbf{x}_i\Vert_2 -\Vert \textbf{x}_j\Vert_2) = \Vert \textbf{x}_i - \textbf{x}_2\Vert_2$\\
\indent Similarly, we can obtain that $d_c^h(\textbf{z}_i,\textbf{z}_j) = \Vert \textbf{x}_i - \textbf{x}_j\Vert_2$ if $\bar{k} = -1$.\\
$\Box$\\
\section{Analysis on the kernel construction by $f(\textbf{x}_i^T\textbf{x}_j)$}
\begin{tcolorbox}
\textbf{Assumption.} Given a non-linearly increasing function $f(a): \mathbb{R}^1 \rightarrow \mathbb{R}^1$, $\kappa(\textbf{x}_i,\textbf{x}_j) = f(\textbf{x}_i^T\textbf{x}_j)$ is a kernel function
\end{tcolorbox}
\textbf{Proof}: Because it is not a strict theorem in our paper, we just give a way to analyze when $f(\textbf{x}_i^T\textbf{x}_j)$ defines a kernel function.\\
\indent We use the Taylor expansion of the $f(a)$, there is
\begin{equation}
    f(a) = f(0) + f^{(1)}(0)a + \frac{f^{(2)}a^2}{2!} +\cdots + \frac{f^{(k)}a^k}{k!} + \cdots
\end{equation}
For easy understanding, we define a function of the matrix $\textbf{A}$ as $ \textbf{Y} = f(\textbf{A})$ as $f(\textbf{A})_{ij} = f(\textbf{A}_{ij})$. Therefore, there exists
\begin{equation}
f(\textbf{x}_i^T\textbf{x}_j) = f(0) + \sum_{k=1}^k f^{(k)}(0)(\textbf{x}_i^T\textbf{x}_j)^k +\cdots.
\end{equation}
 In order to prove $f(\textbf{x}_i^T\textbf{x}_j)$ defines a kernel function, we need to prove that the matrix $\textbf{Y}$ defined by $\textbf{Y}_{ij}=f(\textbf{x}_i^T\textbf{x}_j)$ is a semi-defined matrix.\\
 \indent Because $(\textbf{x}_i^T\textbf{x}_j)^k$ is an $k$-order polynomial kernel function, if we set $k$-order derivative $f^{(k)}(x) \geq 0$ ($k>0$), $f(\textbf{x}_i^T\textbf{x}_j)$ will define a kernel function.\\
 \indent In this paper, we set $f(x) = x^2$ and $f(x) = exp(ax) (a>0)$, whose $k>0$ order derivative is larger than 0, thus, they define kernel functions.\\
$\Box$
\section{Proof of Theorem 2}
\begin{figure}[t]
    \centering
    \footnotesize
    \includegraphics[width=0.9\linewidth]{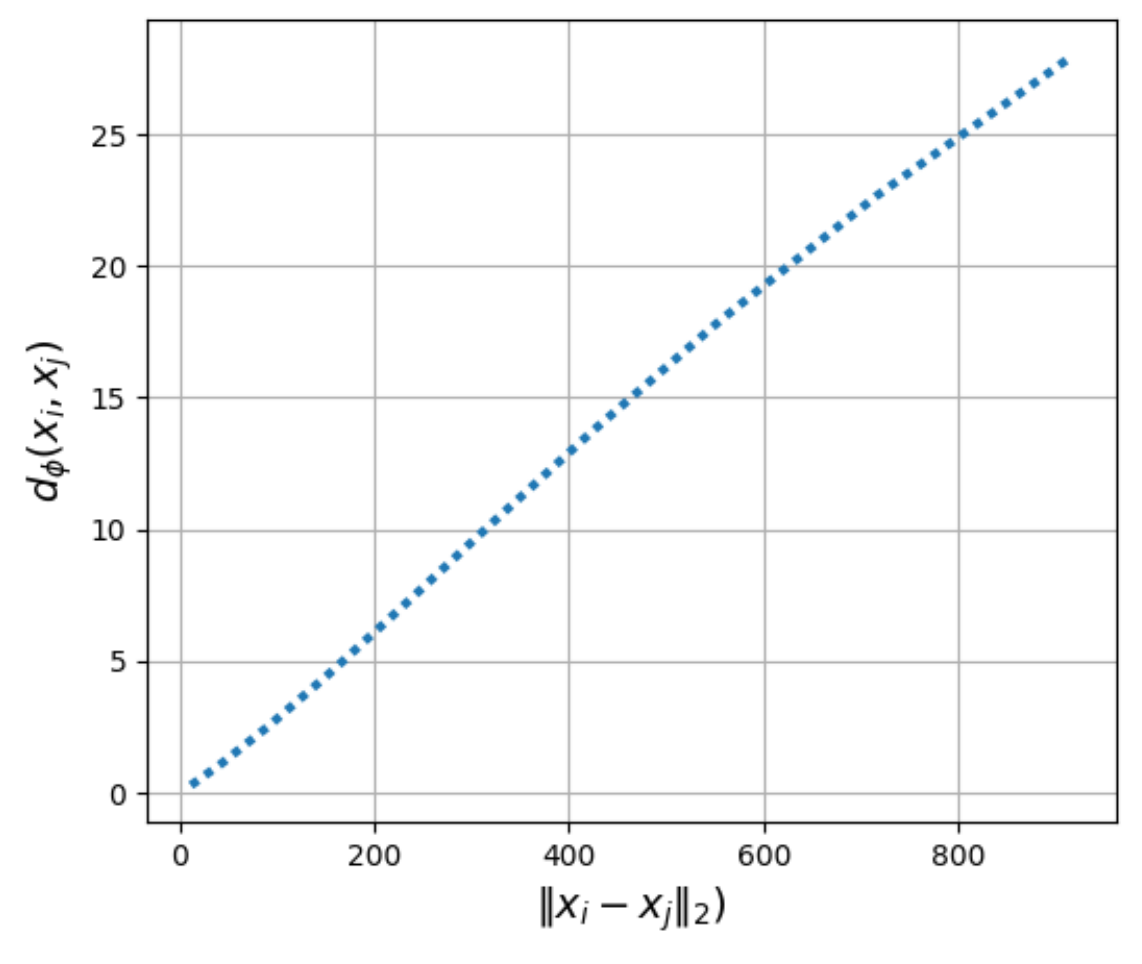}
\caption{\footnotesize{The distance in the kernel space $d_{\phi}(\textbf{x}_i,\textbf{x}_j)$ with non-linearly increase with the distance $\Vert \textbf{x}_i-\textbf{x}_j\Vert_2$}.}
    \label{figure_2_error_4}
\end{figure}

\begin{tcolorbox}
\textbf{Theorem 2.} If $f(a)>0$ is a non-linearly increasing function with respective to $a$, determining a kernel function $\kappa(\textbf{x}_i,\textbf{x}_j) = f(\textbf{x}_i^T\textbf{x}_j)$, there exists another increasing function $g(a)$ to let  $d_{\phi}(\textbf{x}_i,\textbf{x}_j) > g(\Vert \textbf{x}_i - \textbf{x}_j\Vert_2)$.\vspace{-5pt}
\end{tcolorbox}
\textbf{Proof}. Suppose the non-linear projection determined by the kernel function is $\kappa(\textbf{x}_i,\textbf{x}_j) = f(\textbf{x}_i^T\textbf{x}_j)$ as $\textbf{z}_i=\Phi(\textbf{x}_i)$. Thus, $d^2_{\phi}(\textbf{x}_i,\textbf{x}_j)=\Vert \Phi(\textbf{x}_i) - \Phi(\textbf{x}_j)\Vert_2^2 = \kappa(\textbf{x}_i,\textbf{x}_i) + \kappa(\textbf{x}_j,\textbf{x}_j) - 2\kappa(\textbf{x}_i,\textbf{x}_j) = f(\Vert \textbf{x}_i\Vert_2^2) + f(\Vert \textbf{x}_i\Vert_2^2) - 2f( \textbf{x}_i^T\textbf{x}_j)$\\
\indent Obviously, $f(\Vert \textbf{x}_i\Vert_2^2) + f(\Vert \textbf{x}_i\Vert_2^2) - 2f( \textbf{x}_i^T\textbf{x}_j) = f(\Vert \textbf{x}_i\Vert_2^2) - f( \textbf{x}_i^T\textbf{x}_j) + f(\Vert \textbf{x}_i\Vert_2^2)  - f( \textbf{x}_i^T\textbf{x}_j)$. Let us denote $a_1 = \Vert \textbf{x}_i\Vert_2^2$, $a_2 = \Vert \textbf{x}_j\Vert_2^2$, and $a_3 = \textbf{x}_i^T\textbf{x}_j$. Without loss of generality, we set $a_1< a_3$ and $a_3<a_2$. In this way, we can obtain that $d_{\phi}(\textbf{x}_i,\textbf{x}_j) = f'(a_1^*) (a_1 - a_3) + f'(a_2^*)(a_2 - a_3)$ where $a_1^* \in (a_1,a_3)$ and $a_2^*\in (a_3,a_2)$. Because $f(x)$ is an increasing function, $f'(a_1^*) >0 $ and $f'(a_2^*)>0$. In this way, we obtain that $d_{\phi}(\textbf{x}_i,\textbf{x}_j) > f'(a_1^*) (a_1 + a_2-2a_3)$. Because $a_1^* \in (a_1,a_3)$ and $a_2^*\in (a_3,a_2)$, they are the functions of values $\{a_1,a_2,a_3\}$, which determine the distance $\Vert \textbf{x}_i - \textbf{x}_j\Vert_2$. Besides, we know that the larger the distance is, the larger $a_3-a_1$ and $a_2-a_3$ would be. Because $f(\textbf{x}_i^T\textbf{x}_j)$ determines a kernel function, there is $f''(x)>0$. Thus, increasing $a_3-a_1$ and $a_2-a_3$ would also increase $a_1^*$ and $a_2^*$. In this way, we know $a_1^*$ and $a_2^*$ are also functions of the distance $\Vert \textbf{x}_i - \textbf{x}_j\Vert_2$. Therefore, there exists an increasing function to let $a_1^* = p(\Vert \textbf{x}_i-\textbf{x}_j \Vert_2)$. Because $f(\textbf{x}_i^T\textbf{x}_j)$ determines a kernel function, there is $f''(x)>0$, $f'(a_1^*) = f'(p(\Vert \textbf{x}_i - \textbf{x}_j\Vert_2^2))$ is also an increasing function. In this way, we have proved the result: $d_{\phi}(\textbf{x}_i,\textbf{x}_j) > g(\Vert\textbf{x}_i - \textbf{x}_j\Vert_2^2) = f'(p(\Vert \textbf{x}_i - \textbf{x}_j\Vert_2^2))\Vert_2^2\textbf{x}_i - \textbf{x}_j\Vert_2^2$.\\
We also present a graphical proof by setting $f(x) = x^2$ in Figure \ref{figure_2_error_4}.
$\Box$
\\

\section{Proof of Theorem 3}
\begin{tcolorbox}
\textbf{Theorem 3.} If a set of bilinear features are linearly separated, their dimension can be reduced by a linear projection without harming the discriminant ability, and the distances among the obtained low-dimensional features are upper-bounded by a constant.
\end{tcolorbox}
\textbf{Proof:} The bilinear pooling $\textbf{b}_i = vec(\textbf{x}_i^T\textbf{x}_i)$ can be seen as the explicit projection of the polynomial kernel functions because $\textbf{b}_i\textbf{b}_j = (\textbf{x}_i^T\textbf{x}_j)^2$. The linear dimension reduction on bilinear pooling can be formulated as $\hat{\textbf{b}}_i = \textbf{L}^T\textbf{b}_i$, where $\textbf{L}$ is the projection for the dimension reduction.\\
\indent We employ the graph partition model to analyze the dimension reduction of the bilinear features. Considering the bilinear features $\{\textbf{b}_i\}_{i=1}^N$ as a set of $N$ nodes, which can be divided into $T$ groups. We denote those groups as $\{\mathbb{G}_i\}_{i=1}^T$. We construct the similarity of the nodes as $\textbf{K} = \textbf{B}^T\textbf{B} \in \mathbb{R}^{N\times N}$ where $\textbf{B} = [\textbf{b}_1,\textbf{b}_2,\cdots,\textbf{b}_N]$, and the degree matrix is $\textbf{D}\in \mathbb{R}^{N\times N}$ with $\textbf{D}_{ii} = \sum_{j=1}^N\textbf{K}_{ij}$. We suppose $\textbf{y}\in \mathbb{R}^{N\times 1}$ is the embedding vector of the nodes. The objective function based on the normalized cuts for the $2$-graph partition is presented as follows.
\begin{equation}
    \min_{\textbf{y}^T\textbf{y} = \textbf{D}} Tr(\textbf{y}^T \textbf{D}^{-1/2}\textbf{K}\textbf{D}^{-1/2} \textbf{y})
\end{equation}
The above optimization problem has a solution:
\begin{equation}
 y_i = \left\{
    \begin{matrix}
        1, \textbf{x}_i \in \textbf{G}_1\\
        -\sqrt{\frac{\delta_2}{\delta_1}}, \textbf{x}_i \in \textbf{G}_2
    \end{matrix}
    \right.
\end{equation}
For the $T$-way partition, we can employ the one-vs-rest strategy to formulate the problem, i.e., we employ the $\textbf{Y} = [\textbf{y}_1,\textbf{y}_2,\cdots,\textbf{y}_T]$ to indicate the different cluster which is presented as follows:
\begin{equation}
    \min_{\textbf{Y}^T\textbf{Y} = \textbf{D}} Tr(\textbf{Y}^T \textbf{D}^{-1}\textbf{K}\textbf{D}^{-1} \textbf{Y})
\end{equation}
\indent The above solution can be obtained by performing the eigenvalue decomposition on the $\textbf{D}^{-1/2}\textbf{K}\textbf{D}^{-1/2}=\textbf{D}^{-1/2}\textbf{B}^T\textbf{B}\textbf{D}^{-1/2}$. And the $\textbf{Y}$ is the eigenvector of $\textbf{D}^{-1/2}\textbf{K}\textbf{D}^{-1/2}$.\\
\indent Let us consider the diagonal elements in $\textbf{D}$. Because $\textbf{D}_{ii} = \sum_{j=1}^N(\textbf{x}_i^T\textbf{x}_j)^2$, can find that $1/\sqrt{\textbf{D}_{ii}} \approx 1 /\sqrt{\textbf{D}_{kk}}$ if $N$ is large. This is because $D_{ii}$ is increased with the increase of $N$. In this way, we can employ $1/\sqrt{\textbf{D}_{ii}} = t \textbf{I}$. Therefore, the eigenvalue on $\textbf{D}^{-1/2}\textbf{K}\textbf{D}^{-1/2}$ is equivalent to the singular value decomposition (SVD) on $\textbf{B}/\sqrt{t}$. The SVD decomposes $\textbf{B}$ as $\textbf{B} = \textbf{U}\Sigma\textbf{V}$ where $\textbf{U}$ and $\textbf{V}$ are the singular vectors and $\Sigma$ stores the singular values. And $\textbf{U}$ is the eigenvector of $\textbf{D}^{-1/2}\textbf{K}\textbf{D}^{-1/2}$.\\
\indent Because $\textbf{V}^T\textbf{V}=\textbf{I}$, we can find that $\textbf{U} = \textbf{D}^{1/2}\textbf{B}\textbf{V}^T\sigma^{-1}$.Therefore, $\textbf{U} = t\textbf{I}\textbf{B}\textbf{V}^T\sigma^{-1} = \textbf{B} t\textbf{V}^T\sigma^{-1}$. In summary, we can employ a projection $\textbf{L}= t\textbf{V}^T\sigma^{-1}$ to reduce the dimension of the bilinear features as $\hat{\textbf{b}} = \textbf{L}^T \textbf{b}$. For the whole dataset, there is $\hat{\textbf{B}} = \textbf{L}^T\textbf{B}$.\\
\indent Because the $\kappa(\textbf{x}_i,\textbf{x}_j) = \textbf{b}_i^T\textbf{b}$, the above dimension reduction is equivalent to performing the linear dimension reduction in the kernel space of polynomial kernel function. Because the procedure is based on the singular value decomposition, which is also adopted by the principal component analysis, the above procedure is similar to performing the kernel principal component analysis on the samples $\{\textbf{x}_i\}_{i=1}^N$.\\
\indent Because the norm of any column in $\hat{\textbf{B}}\in \mathbb{R}^{N\times T}$ is smaller than 1, the distance between any two samples can be bounded by $\sqrt{(2)^2 + 2^2} = 2\sqrt{2}$.\\

\section{Results on additional data-splits}
\label{results}
Table~\ref{table_10} extends additional results on the Full split on the citation datasets as introduced in~\cite{chen2018fastgcn}.

\begin{table}[h]
\centering
\scriptsize
\setlength{\tabcolsep}{4pt}
\caption{Results on the Full split. \textsc{auroc} for the Link Prediction (\textsc{LP}) task and \textsc{accuracy}(\%) in the multi-class Node Classification (\textsc{NC}) task. Results are provided for 3 random seeds.}
    \begin{tabular}{lccc}
        \toprule
         {\textsc{Full-Split}}& {\begin{tabular}{c}\textbf{PubMed} \\($\delta = 3.5$)\end{tabular}}  & {\begin{tabular}{c}\textbf{CiteSeer} \\($\delta = 5$)\end{tabular}} & {\begin{tabular}{c}\textbf{Cora} \\($\delta =11$)\end{tabular}}\\
        \midrule
        
        GCN              &  87.9$\pm$0.5   &  78.8$\pm$1.0  &   87.1$\pm$0.5  \\
        GCNII            &  85.6$\pm$0.6    &  77.1$\pm$0.2  &  82.1$\pm$3.1  \\
        GAT              &  84.4$\pm$0.3    &  76.6$\pm$0.3  &   83.0$\pm$0.6 \\
        GraphSAGE        &  88.1$\pm$0.3    &  79.2$\pm$0.5  &   87.7$\pm$0.3  \\
        TransformerConv  &  \textbf{88.7$\pm$0.4 } & 79.5$\pm$0.4  &    87.0$\pm$0.4  \\
        \arrayrulecolor{black!30}\cmidrule{1-4}
        BP (Ours)        &  87.5$\pm$0.2  &   76.2$\pm$0.6  &    87.7$\pm$0.9  \\
        HBP (Ours)       &  87.4$\pm$0.2  &   78.8$\pm$1.0  &    87.8$\pm$0.3  \\
        HBP-$\mathbb{KR}$ (Ours)     &  87.6$\pm$0.1  & \textbf{ 79.2$\pm$0.4}  &\textbf{  89.3$\pm$0.5}  \\
        
        \arrayrulecolor{black}\bottomrule
    \end{tabular}
\label{table_10}
\end{table}

\begin{table*}[h]
\centering
\footnotesize
\setlength{\tabcolsep}{5pt}
\renewcommand{\arraystretch}{1.1}
\caption{Hyperparameter settings for the GNN models.}
  \begin{tabular}{@{}lcccc@{}}
    \arrayrulecolor{black}\toprule
    Model & Hidden Channels & Layers & Heads & Additional Hyperparameters \\
    \midrule
    GCN & 16 & 2 & - & - \\
    GAT & 16 & 2 & 8 & - \\
    Transformer & 16 & 2 & 2 & - \\
    GCNII & 64 & 64 & - & $\alpha=0.1, \theta=0.5$ (as specified in original paper) \\
    \arrayrulecolor{black!30}\cmidrule{1-5}
    SAGE (OGB) &  128  &  3  & -  &  - \\
    GAT (OGB) & 128 & 3 & 4 & Skip connections between layers \\
    \arrayrulecolor{black!30}\cmidrule{1-5}
    HGCN & 16 & 2 & - & - \\
    \arrayrulecolor{black!30}\cmidrule{1-5}
    HBP-GCNII & 64 & 64 & - & $c=1$\\
    HLBP-GCNII & 64 & 64 & - & $c=1$, low-rank dimension$=16$, low-rank heads$=3$\\
    \arrayrulecolor{black}\bottomrule
  \end{tabular}
\label{table_6}
\end{table*}

\section{Hyperparameter setting}
\label{hyperparameter}
Our experiments were designed maintaining a consistent set of hyperparameters wherever feasible. Specifically, for the Euclidean-based models GCN and GAT, we set the architecture to have 16 hidden channels and a depth of 2 layers as previously reported hyperparameters in the literature for the citations datasets~\cite{kipf2016semi}. The GAT model was further specified to utilize 8 attention heads whereas in the transformer convolutions, the number of heads was set to 2, which was determined to be optimal for the performance metrics of interest~\cite{velickovic2017graph}. For the GCNII model, we adhered to the hyperparameter settings described in the original publication~\cite{chen2020simple}, which uses 64 hidden channels and layers. Regarding the models used in OGB, we keep conservative hyperparameters, with 128-dimension embedding in 3 layers and 4 heads in the case of GAT. A summary is presented in Table~\ref{table_6}. Across all model configurations, we implemented a uniform dropout rate of 0.6 to mitigate overfitting and maintained consistent optimizer parameters, utilizing the Adam optimizer with a learning rate of 0.01 in the small datasets and 0.001 in OGB. This uniform approach to the experimental setup was aimed at providing a fair evaluation of the methods.

\begin{figure}[t]
    \centering
    \footnotesize
        
    \subfigure[]{\includegraphics[width=0.49\linewidth]{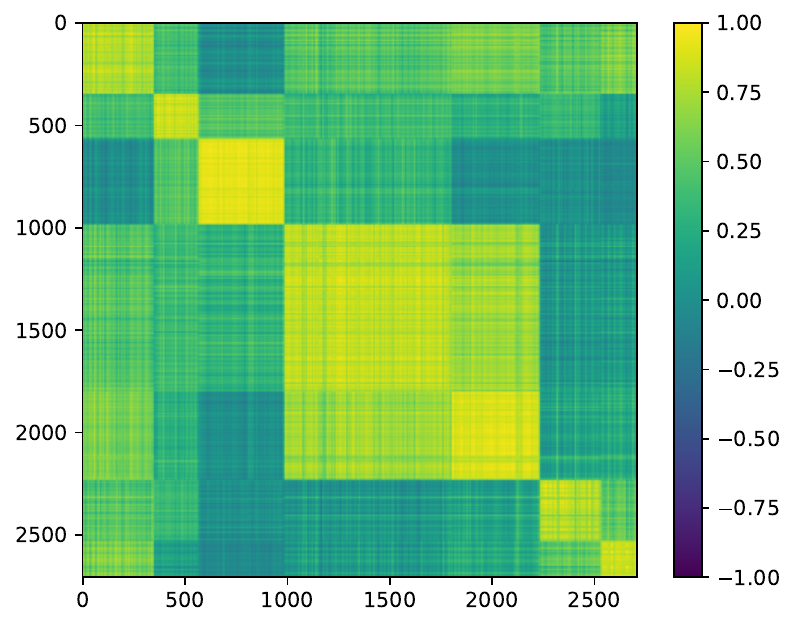}}
    \subfigure[]{\includegraphics[width=0.49\linewidth]{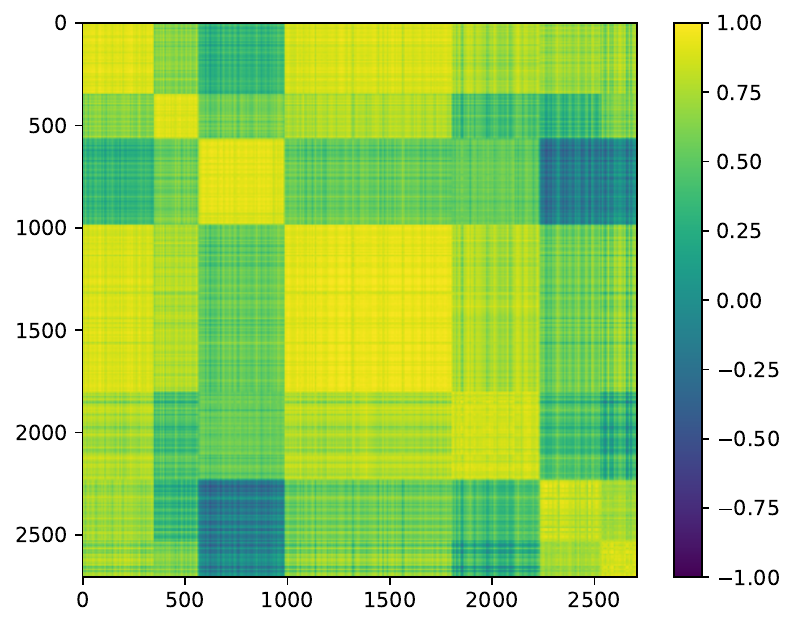}} 
\caption{Cosine similarity between the learned embeddings for GCNII on (a) the Euclidean space and (b) the hyperbolic HBP alternative in the Cora dataset. Whereas a discriminant ability in the Euclidean model comes mostly from angular differentiation, in the case of the hyperbolic alternative, this is significantly less pronounced. }
    \label{figure_2}
\end{figure}

\section{Related Work}

\textbf{Bilinear Pooling.} Bilinear pooling was initially used to enhance the discriminative ability of visual features in a bag-of-word approach~\cite{koniusz2016higher}. Later, it was used to significantly improve fine-grained classification in an end-to-end learning framework~\cite{lin2017bilinear}. Subsequently, numerous studies were conducted on bilinear pooling and similar methods for various visual tasks \cite{gao2016compact,gao2020revisiting,kong2017low}. These include designing novel bilinear pooling, normalization strategies, and studying dimensionality reduction techniques. Beyond image recognition tasks, research expanded into graph-structured data. For instance, some studies investigated the use of bilinear pooling for drug discovery~\cite{bai2023interpretable}. Other research showed that bilinear pooling could enhance features of graph neural networks on graph classification, as it obtains properties like invariance \cite{wang2020second}. However, whether it is on fine-grained data or graph-structured data, few studies suggest that bilinear pooling can be used to enhance hierarchical representations.

\textbf{Hyperbolic Representation Learning.} Hyperbolic representation learning on the other side was first applied in natural language processing. It uses the distortion-free properties of hyperbolic space to approximate tree structures and learn vectors representing hierarchical category information. To effectively utilize the properties of hyperbolic space, various learning algorithms were developed, such as hyperbolic space-based kernel methods~\cite{fang2021kernel}, hyperbolic neural networks~\cite{chami2019hyperbolic}, hyperbolic space metric learning~\cite{vinh2020hyperml}, hyperbolic discriminative classifiers~\cite{cho2019large}, among others. Later, to enhance the performance of hyperbolic space representation learning, regularization methods such as alignment and curvature learning algorithms~\cite{yang2023hyperbolic}, numerical stability regularization~\cite{mishne2023numerical}, and norm-wrap-based gradient updating methods~\cite{nagano2019wrapped} were proposed. However, these methods assume that projecting features into hyperbolic space automatically leads to learning hierarchical information. In contrast, this paper focuses on enhancing the ability of hyperbolic space learning to capture hierarchical features.

\textbf{Graph Neural Networks.} On the other side, the field of graph learning has witnessed significant research interest and GNNs have become a standard in many machine learning tasks. Graph Convolution Networks (GCN)~\cite{kipf2016semi} are one type of powerful model for representing graph-structured data. Despite their success, GCNs are inherently shallow due to the over-smoothing problem~\cite{keriven2022not}. Although some approaches as GCNII~\cite{chen2020simple} have extended the learning ability enabling deeper models, or powerful attention-based feature extractors~\cite{zhang2021hyperbolic, shi2020masked} have been embedded into the message-passing strategy, effectively inducing the right representation remains an open challenge. 

In that sense, several works in the literature have used hyperbolic learning to enhance the representation learned by the GNN. However, this has not always translated into beneficial models from the performance and stability perspective. E.g., \cite{chami2019hyperbolic} discusses the sensitivity issues related to learning the curvature in the intermediate layers. Also, simple operations required in message passing and aggregation steps require in some cases to operate back in the tangent space, \cite{dai2021hyperbolic}. Our approach differs from the mentioned ones in the sense that we can directly adopt the potential of Euclidean GNNs, i.e., our method is directly applicable to competitive architectures, and bringing this architecture into the hyperbolic space using a simple projection. Lastly, differently from \cite{chami2019hyperbolic}, that does not report benefit in classifying in the hyperbolic space, and requires mapping the features to the tangent space.

\end{document}